\documentclass[10pt,twocolumn,letterpaper]{article}

\usepackage[pagenumbers]{cvpr} % To force page numbers, e.g. for an arXiv version

\definecolor{cvprblue}{rgb}{0.21,0.49,0.74}
\usepackage[pagebackref,breaklinks,colorlinks,allcolors=cvprblue]{hyperref}
\usepackage{multirow}
\usepackage{array}
\usepackage{graphicx}
\usepackage[american]{babel}
\usepackage{microtype}
\def\paperID{12932} % *** Enter the Paper ID here
\def\confName{CVPR}
\def\confYear{2026}

\title{Continuous Exposure-Time Modeling \\
for Realistic Atmospheric Turbulence Synthesis}
\vspace{-.5cm}
\author{Junwei Zeng$^{1}$, Dong Liang$^{1}$\thanks{Corresponding author.}, Sheng-Jun Huang$^{1}$, Kun Zhan$^{2}$, Songcan Chen$^{1}$\\
\small{\textsuperscript{1}  Nanjing University of Aeronautics and Astronautics}\\
\small{\textsuperscript{2} School of Information Science and Engineering, Lanzhou University}\\
{\tt\small \{zengjunwei,liangdong,huangsj,chensongcan\}@nuaa.edu.cn, kzhan@lzu.edu.cn }
}

\begin{document}

\maketitle

\begin{abstract}
Atmospheric turbulence significantly degrades long-range imaging by introducing geometric warping and exposure-time-dependent blur, which adversely affects both visual quality and the performance of high-level vision tasks. Existing methods for synthesizing turbulence effects often oversimplify the relationship between blur and exposure-time, typically assuming fixed or binary exposure settings. This leads to unrealistic synthetic data and limited generalization capability of trained models. To address this gap, we revisit the modulation transfer function (MTF) formulation and propose a novel Exposure-Time-dependent MTF (ET-MTF) that models blur as a continuous function of exposure-time. For blur synthesis, we derive a tilt-invariant point spread function (PSF) from the ET-MTF, which, when integrated with a spatially varying blur-width field, provides a comprehensive and physically accurate characterization of turbulence-induced blur. Building on this synthesis pipeline, we construct ET-Turb, a large-scale synthetic turbulence dataset that explicitly incorporates continuous exposure-time modeling across diverse optical and atmospheric conditions. The dataset comprises 5,083 videos (2,005,835 frames), partitioned into 3,988 training and 1,095 test videos. Extensive experiments demonstrate that models trained on ET-Turb produce more realistic restorations and achieve superior generalization on real-world turbulence data compared to those trained on other datasets. The dataset is publicly available at: \url{github.com/Jun-Wei-Zeng/ET-Turb}.
\end{abstract}   
\section{Introduction}
\label{sec:intro}

\begin{figure}[htbp]
  \centering
  \includegraphics[width=\columnwidth]{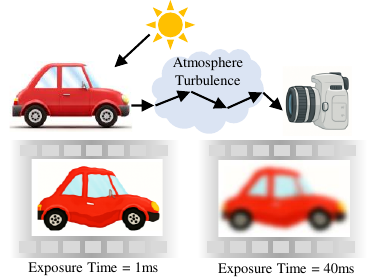} 
  \caption{Imaging results under atmospheric turbulence with different exposure-time. \textbf{Short-exposure} (e.g., 1 ms) primarily exhibits turbulence-induced \textbf{tilt} as turbulence state is effectively frozen. In contrast, \textbf{long-exposure} (e.g., 40 ms) integrates multiple turbulent states over time, resulting in significantly stronger \textbf{blur}.}
  \label{fig:camera} 
  \vspace{-.5cm}
\end{figure}

Atmospheric turbulence, caused by random fluctuations in the refractive index of air~\cite{fried1966optical,tatarski2016wave}, distorts optical wavefronts and introduces both geometric warping and exposure-time-dependent blur in long-range imaging systems~\cite{chan2022tilt}. The spatio-temporal randomness of turbulence poses significant challenges for data-driven restoration methods, limiting their ability to accurately model and mitigate such degradations. These effects severely impact a wide range of applications, including remote sensing~\cite{xiang2020coherent}, video surveillance~\cite{mei2023ltt}, object recognition\cite{shi2018aster,yasarla2021learning}, and astronomical observation~\cite{ma2020night}.

Although learning-based approaches for turbulence mitigation~\cite{chak2018subsampled,lau2019restoration,zhu2012removing,jaiswal2023physics,lau2021atfacegan,mao2022single,nair2021confidence,nair2023ddpm,anantrasirichai2023atmospheric,jin2021neutralizing,zhang2024imaging,zhang2025learning} have demonstrated considerable progress, their performance remains dependent on the realism and diversity of training data. Acquiring large-scale paired real-world turbulence data is both expensive and impractical, making high-quality synthetic datasets an essential alternative for training and evaluation.

As illustrated in Fig.~\ref{fig:camera}, the appearance of turbulence-induced blur varies substantially with exposure-time. Short-exposure images predominantly exhibit geometric distortions (tilt), while long-exposure images accumulate more blur due to temporal integration of turbulent states. Despite this physical relationship, most existing synthetic turbulence generation methods overlook exposure-time as a continuous variable. Numerous approaches~\cite{anantrasirichai2013atmospheric,gilles2017open,hirsch2010efficient,xu2024long,jin2021neutralizing,boehrer2021turbulence} employ a fixed exposure setting across all samples, resulting in images with similar blur statistics that fail to capture the temporal variability observed in real imaging. Consequently, models trained on such data exhibit limited generalization when applied to real-world turbulence.

Some methods have attempted to incorporate exposure-time effects but treat them in a simplified, binary manner, categorizing exposures only as ``short'' or ``long'' and applying the corresponding atmospheric modulation transfer function (MTF) for simulation~\cite{mao2021accelerating,chimitt2022real,chimitt2020simulating,fried1978probability,carbillet2013astronomical}. These coarse approximations ignore the nature of exposure-time in real cameras, where intermediate values produce gradual and physically meaningful transitions in blur characteristics~\cite{du2015detector,azoulay1990effects}. As a result, synthesized imagery exhibits abrupt changes between exposure regimes rather than the smooth blur dynamics present in actual turbulent conditions.

The fundamental limitation of existing methods stems from discrete MTF modeling, which governs turbulence-induced blur in the frequency domain~\cite{williams2002introduction}. To address this gap, we introduce an Exposure-Time-dependent MTF (ET-MTF) that characterizes blur as a continuous function of exposure-time. Building on the decomposition of turbulence degradation into tilt and blur components~\cite{chan2022tilt,charnotskii2022warp}, we derive a tilt-invariant point spread function (PSF) from the ET-MTF to generate pure blur effects. The PSF is then combined with a spatially varying blur-width field to jointly capture the temporal and spatial characteristics of turbulence-induced blur. The integration of these components forms a physically-grounded turbulence synthesis pipeline, which we use to construct ET-Turb, a large-scale synthetic turbulence dataset that explicitly incorporates continuous exposure-time variation. Extensive experiments demonstrate that our approach produces more realistic turbulence simulations and significantly improves the generalization performance of trained models on real-world data.

Our main contributions are summarized as follows:
\begin{itemize}
  \item We propose a physics-inspired turbulence synthesis pipeline that continuously models the relationship between exposure-time and turbulence-induced blur, overcoming the limitations of prior discrete or binary exposure modeling approaches.
  
  \item We derive an Exposure-Time-dependent MTF (ET-MTF) that enables smooth interpolation across the full exposure spectrum, from which we obtain a tilt-invariant PSF combined with a spatially varying blur-width field for unified and physically accurate turbulence blur synthesis.
  
  \item We construct the ET-Turb dataset using this pipeline, explicitly incorporating exposure-time as a continuous variable throughout the synthesis process to generate more realistic turbulence data and enhance model generalization across diverse imaging conditions.
\end{itemize}

% \vspace{-.2cm}
\section{Related Work}
\label{sec:related_work}

Acquiring paired clear and turbulent images from real-world scenarios for training deep learning-based turbulence mitigation models remains a significant challenge. Consequently, the research community has increasingly relied on synthetic turbulence datasets to facilitate model development. In this section, we provide a comprehensive review and critical analysis of existing turbulent image simulation techniques, categorizing them into three main paradigms.

\textbf{Physical Turbulence Simulation.}  
Many methods employ physical apparatus such as gas hobs or heat sources to generate controllable indoor turbulence environments~\cite{jin2021neutralizing,carbillet2013astronomical,mao2021accelerating}. While these setups offer intuitive and direct simulation of turbulence effects, they are typically constrained to short optical paths. This limitation results in oversimplified turbulence statistics~\cite{hill2025deep} that fail to capture the multi-scale dynamics characteristic of real atmospheric conditions. Moreover, the inherent restrictions in hardware configurability limit the diversity of generated turbulence patterns, consequently reducing the transferability of models trained on such data to practical outdoor imaging scenarios.

\textbf{Multi-step Phase Screen Simulation.}  
Another prominent approach discretizes the turbulent medium into a sequence of spatially separated phase screens to model wavefront distortions over long propagation paths~\cite{giles2000setting,roggemann1995method,bos2012technique,schmidt2010numerical}. This methodology achieves high physical fidelity by explicitly simulating light propagation through turbulent layers. However, it comes at the cost of substantial computational complexity, typically requiring hundreds of phase screens for accurate simulation, which imposes significant computational and memory overhead~\cite{hardie2017simulation}. Consequently, the prohibitive resource demands render large-scale dataset generation based on this approach economically impractical.

\textbf{Zernike Polynomial Modeling.}  
To address the computational inefficiency of phase screen methods, several studies have adopted Zernike polynomials to approximate phase aberrations~\cite{takato1995spatial,noll1976zernike,chimitt2020simulating}. Notably, TurbSim~\cite{chimitt2020simulating} simplifies the propagation process by collapsing it into a single equivalent phase screen with statistically correlated Zernike coefficients. Subsequent works, such as P2S~\cite{mao2021accelerating}, further accelerate simulation through the use of multilayer perceptrons (MLPs), while DF-P2S~\cite{chimitt2022real} enhances accuracy by refining the correlation matrix formulation. Despite these advancements, these models remain fundamentally constrained to binary exposure regimes, typically supporting only short or long-exposure conditions~\cite{chimitt2022real,zhang2024spatio,mao2021accelerating,chimitt2020simulating}. The absence of explicit modeling for intermediate exposure-time significantly limits their realism and practical applicability in comprehensive atmospheric turbulence synthesis.

Despite substantial advancements in turbulence simulation methodologies, existing approaches predominantly overlook exposure-time as a continuous variable, even though it plays a critical role in shaping turbulence-induced blur. The absence of explicit exposure-time modeling results in synthetic imagery that fails to reproduce the gradual blur transitions observed in real imaging systems. In contrast, our method introduces a novel physical turbulence synthesis pipeline that directly embeds exposure-time as a continuous parameter throughout the degradation process. This approach enables the generation of exposure-aware synthetic data that more accurately captures real-world turbulence phenomena, thereby providing superior training and evaluation resources for turbulence mitigation.
% \vspace{-.2cm}
\section{Atmospheric Turbulence Synthesis}
\label{sec:method}

\subsection{Preliminaries}\label{subsec:imaging_model}
The formation of a turbulence-degraded image can be mathematically modeled as~\cite{shimizu2008super,leonard2012simulation,chan2022tilt,charnotskii2022warp,lau2019restoration}:
\begin{equation}
I(\mathbf{x}) = \mathcal{H}(J(\mathbf{x})) = \mathcal{B}(\mathcal{T}(J(\mathbf{x})))
\label{eq:imaging_model}
\end{equation}
where $J(\mathbf{x})$ denotes the clean image at spatial position $\mathbf{x} \in \mathbb{R}^2$, and $\mathcal{H}(\cdot)$ represents the overall turbulence degradation operator. This degradation process can be decomposed into two primary components: geometric tilt $\mathcal{T}(\cdot)$ and blur $\mathcal{B}(\cdot)$.

\noindent\textbf{Tilt.}
Tilt-induced degradation manifests as geometric distortion in the image plane, resulting from low-order wavefront aberrations. The statistical characteristics of this geometric warping exhibit negligible dependence on exposure-time. We adopt the stochastic modeling framework proposed by Schwartzman et al.~\cite{schwartzman2017turbulence}, representing tilt as spatially correlated two-dimensional displacement fields.

\noindent\textbf{Blur.}
Turbulence-induced blur is mathematically described as a convolution operation~\cite{miller2019data}:
\begin{equation}
\mathcal{B}(I_T(\mathbf{x})) = \text{PSF}(\mathbf{x}) \ast I_T(\mathbf{x})
\label{eq:psf_blur}
\end{equation}
where $I_T(\mathbf{x}) = \mathcal{T}(J(\mathbf{x}))$ represents the tilt image, PSF denotes the point spread function characterizing the blur kernel, and $\ast$ denotes the convolution operator.

In the frequency domain, the PSF corresponds to the optical transfer function (OTF), defined as the complex-valued Fourier transform of the PSF:
\begin{equation}
\text{OTF}(\boldsymbol{\xi}) = \mathcal{F}\{\text{PSF}(\mathbf{x})\} = \text{MTF}(\boldsymbol{\xi}) \, e^{i\cdot\text{PhTF}(\boldsymbol{\xi})}
\label{eq:otf_def}
\end{equation}
where $\text{MTF}(\boldsymbol{\xi})$ denotes the modulation transfer function encoding the blur effect, and $\text{PhTF}(\boldsymbol{\xi})$ denotes the phase transfer function encoding the tilt effect. $\boldsymbol{\xi}$ represents the spatial frequency, with high frequencies corresponding to fine image details such as edges, and low frequencies representing coarse structures and overall scene composition.

To model pure blur, we focus solely on the MTF component~\cite{miller2019data}. Therefore, Eq.~\eqref{eq:psf_blur} can be expressed in the frequency domain as:
\begin{equation}
\mathcal{B}(I_T(\boldsymbol{\xi})) = \text{MTF}(\boldsymbol{\xi}) \cdot I_T(\boldsymbol{\xi})\,.
\label{eq:mtf_blur}
\end{equation}

The magnitude of $\text{MTF}(\boldsymbol{\xi})$ directly dictates the extent of blur at each spatial frequency. When $\text{MTF}(\boldsymbol{\xi})$ is high, the corresponding frequency content is well preserved with weak blur. In contrast, when $\text{MTF}(\boldsymbol{\xi})$ is low, the frequency experiences substantial attenuation, leading to strong blur and the loss of fine details.

Several studies~\cite{fried1966optical, krapels2001atmospheric} have explored the design of the MTF for atmospheric turbulence, providing mathematical expressions for MTF in both short and long-exposure regimes.

In the short-exposure regime, the exposure-time is brief enough for atmospheric turbulence to remain frozen, resulting in an instantaneous turbulence realization recorded by the sensor, with images dominated by tilt and weak blur. The short-exposure MTF~\cite{krapels2001atmospheric} is denoted as $\text{MTF}_{\text{SE}}$:
\begin{equation}
\text{MTF}_{\text{SE}}(\boldsymbol{\xi}) = e^{-57.4 a C_n^2 L \lambda^{-1/3} \|\boldsymbol{\xi}\|^{5/3} \left[1 - \mu \left( \left( \frac{\lambda\|\boldsymbol{\xi}\|}{D} \right)^{1/3} \right) \right]}
\end{equation}
where $a$ is a wave shape constant of 3/8 for spherical and 1 for plane waves, $C_n^2$ denotes the refractive index structure parameter, $L$ is the propagation distance, $\lambda$ is the wavelength, $\mu$ is 0.5 in the far field and 1 in the near field, and $D$ is the aperture diameter. The factor $1 - \mu\bigl(\frac{\lambda\|\boldsymbol{\xi}\|}{D}\bigr)^{1/3}$ acts to slow the decay of $\text{MTF}_{\text{SE}}$. As $\|\boldsymbol{\xi}\|$ increases, this term prevents $\text{MTF}_{\text{SE}}$ from rapidly approaching zero, maintaining higher values at fine spatial scales and preserving details.

In the long-exposure regime, the sensor integrates multiple rapidly varying turbulent states over time, leading to the progressive accumulation of strong blur. The corresponding MTF~\cite{krapels2001atmospheric}, denoted as $\text{MTF}_{\text{LE}}$, can be given as:
\begin{equation}
\text{MTF}_{\text{LE}}(\boldsymbol{\xi}) = e^{-57.4 a C_n^2 L \lambda^{-1/3} \|\boldsymbol{\xi}\|^{5/3} }\,.
\end{equation}

In contrast to $\text{MTF}_{\text{SE}}$, this formulation excludes the component that mitigates high-frequency attenuation.  As the spatial frequency increases, $\text{MTF}_{\text{LE}}$ decays rapidly toward zero, resulting in a loss of fine details, which aligns with the theoretical expectation that longer exposure-time leads to more pronounced image blurring.

While these formulations accurately characterize the two asymptotic regimes, it does not specify how image blur evolves continuously with exposure-time. Existing works~\cite{beaumont1996experimental,rasouli2006measurement} address this gap by empirically interpolating between $\text{MTF}_{\text{SE}}$ and $\text{MTF}_{\text{LE}}$, providing approximate models for the transition between the short and long-exposure regimes. However, these approaches do not explicitly model the exposure-dependent integration of turbulence realizations, which limits their physical interpretability and generalization to different imaging configurations.

\begin{figure*}[t]
\centering
\includegraphics[width=0.9\textwidth]{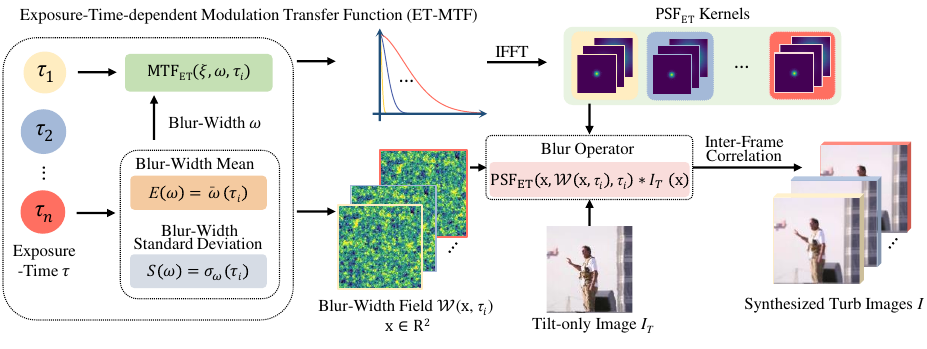}
\caption{Overview of the proposed exposure-time-dependent turbulence synthesis pipeline. The exposure-time $\tau$ is treated as a continuous input parameter to both the ET-MTF and blur-width field formulations. By systematically varying $\tau$, our pipeline generates turbulence-degraded images with physically consistent blur characteristics from the input tilt image.}
\label{fig:pipeline_overview}
\end{figure*}
To address this fundamental limitation, we propose an exposure-time-parameterized imaging model:
\begin{equation}
I(\mathbf{x}) = \mathcal{B}_\tau(\mathcal{T}(J(\mathbf{x})))
\label{eq:imaging_model_tau}
\end{equation}
where the blur operator $\mathcal{B}_\tau(\cdot)$ explicitly incorporates exposure-time $\tau$ as a continuous parameter, enabling physically consistent modeling of blur across the entire exposure-time spectrum. Our exposure-time-dependent turbulence synthesis pipeline is illustrated in Fig.~\ref{fig:pipeline_overview}.

The construction of the exposure-time-dependent blur operator $\mathcal{B}_\tau(\cdot)$ proceeds through three systematic stages. \S\ref{subsubsec:theory_to_etmtf} derives the Exposure-Time-dependent MTF (ET-MTF) from Azoulay's theory~\cite{azoulay1990effects}, providing a physically motivated MTF that smoothly evolves between the short and long-exposure regimes. \S\ref{subsubsec:etmtf_to_psf} transforms this frequency-domain formulation into a spatial-domain PSF suitable for pure blur synthesis. \S\ref{subsubsec:blur_field} extends the scalar blur-width parameter to a spatially varying field formulation constrained by optical turbulence statistics, enabling spatially heterogeneous blur modeling while maintaining temporal consistency.

\subsection{Exposure-Time-dependent Blur Operator}
\subsubsection{From Azoulay's Theory to ET-MTF}\label{subsubsec:theory_to_etmtf}
To model blur degradation across the full spectrum of exposure-time, we extend the finite exposure MTF theory originally proposed by Azoulay~\cite{azoulay1990effects}. Azoulay posits that under short-exposure conditions, the sensor captures a turbulence snapshot with eddies effectively frozen within the aperture diameter $D$. Under long-exposure conditions, the sensor integrates successive turbulent states over time, approximating turbulence as frozen within a larger effective aperture diameter. By varying this effective aperture diameter, MTFs corresponding to different exposure-time can be generated. Following Azoulay’s finite-exposure MTF model~\cite{azoulay1990effects}, we adopt the following Exposure-Time-dependent MTF:
\begin{equation}
\text{MTF}_{\text{ET}}(\boldsymbol{\xi}, \tau) = 
e^{-\left(\frac{\lambda \|\boldsymbol{\xi}\|}{\rho_p(\tau)}\right)^{5/3}}
\label{eq:mtf_et}
\end{equation}
where $\rho_p(\tau)$ denotes the effective coherence length, a concept that connects the aperture diameter $D$ and exposure-time $\tau$, characterizing how the spatial extent of turbulence contributing to blur evolves as exposure-time increases. The relationship between the effective coherence length $\rho_p(\tau)$ and the aperture diameter $D$ is given by:
\begin{equation}
\rho_p(\tau) = 1 + 0.35\left(\frac{r_0}{D + v_w \tau}\right)^{1/3}
\label{eq:rho_p}
\end{equation}
where $r_0$ is the Fried parameter~\cite{fried1966optical} and $v_w$ is the wind velocity. The effective aperture term $D + v_w \tau$ encodes the temporal evolution of blur: as $\tau$ increases, the effective aperture smoothly grows beyond the physical diameter $D$, capturing the transition from short to long-exposure behavior.

While Eq.~\eqref{eq:mtf_et} successfully establishes continuous temporal dependence, it produces spatially uniform blur since $\rho_p(\tau)$ remains constant across the image plane. Real atmospheric turbulence, however, exhibits spatially varying blur patterns due to local refractive index fluctuations. To incorporate this spatial heterogeneity, we reparameterize the ET-MTF in terms of blur-width.

The blur-width $\omega$ is obtained from the full width at half maximum (FWHM) of PSF~\cite{campbell2009atmospheric}, which describes the actual size of the blur spot formed on the image plane by an ideal point source in the imaging system. Geometrically, it represents the diameter at which the PSF intensity drops to half of its peak value, thereby providing an intuitive measure of local blur severity in the spatial domain. It is given by:
\begin{equation}
\omega \approx \frac{0.49 \lambda f}{r_0}
\label{eq:blur_width}
\end{equation}
where $f$ denotes focal length. Inverting this relationship and substituting into Eq.~\eqref{eq:rho_p} yields:
\begin{equation}
\rho_p(\omega, \tau) = 1 + 0.28\left(\frac{\lambda f}{\omega (D + v_w \tau)}\right)^{1/3}.
\label{eq:rho_p_omega}
\end{equation}

The coherence length, previously controlled solely by exposure-time, now explicitly depends on the combined effect of local blur-width and exposure-time. This reparameterization preserves the effective aperture term $D + v_w \tau$ governing temporal behavior, while introducing spatial variability through the position-dependent blur-width.

\begin{figure}[t]
  \centering
  \begin{minipage}[t]{0.9\linewidth}
    \centering
    \includegraphics[width=\linewidth]{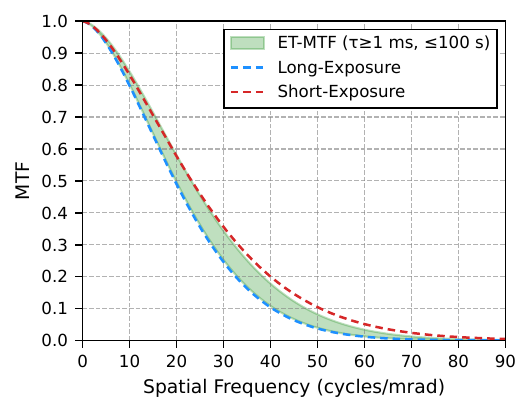}
  \end{minipage}
  \caption{Comparison of MTF formulations at short and long-exposure limits. ET-MTF shows a smooth evolution of the MTF as exposure-time varies between the short and long-exposure limits.}
  \label{fig:mtf_exposure}
\end{figure}

Combining Eq.~\eqref{eq:rho_p_omega} with Eq.~\eqref{eq:mtf_et} gives the final form:
\begin{equation}
\text{MTF}_{\text{ET}}(\boldsymbol{\xi}, \omega, \tau) =
e^{-\left(\frac{\lambda \|\boldsymbol{\xi}\|}{\rho_p(\omega, \tau)}\right)^{5/3}}
\label{eq:final_mtf}
\end{equation}
where $\rho_p(\omega, \tau)$ is defined by Eq.~\eqref{eq:rho_p_omega}, and Fig.~\ref{fig:mtf_exposure} illustrates how the proposed ET-MTF smoothly evolves between the short and long-exposure regimes via exposure-time $\tau$.

To synthesize turbulence-induced blur in the spatial domain while isolating pure blur from tilt, we derive a suitable PSF from the ET-MTF, as detailed in the next subsection.

\subsubsection{From ET-MTF to PSF}
\label{subsubsec:etmtf_to_psf}

Since tilt and blur are modeled separately in our framework, the blur-inducing PSF depends only on the ET-MTF and remains independent of phase terms. Applying the inverse Fourier transform to the ET-MTF yields a tilt-invariant PSF representing pure exposure-time-dependent blur:
\begin{equation}
\text{PSF}_{\text{ET}}(\mathbf{x}, \omega, \tau) =
\mathcal{F}^{-1}\left\{\,\text{MTF}_{\text{ET}}(\boldsymbol{\xi}, \omega, \tau)\,\right\}.
\label{eq:psf_et}
\end{equation}

This PSF characterizes the blur kernel as a function of both local blur-width $\omega$ and global exposure-time $\tau$, without interference from tilt-related phase distortions.

In real turbulent conditions, blur intensity exhibits spatial variation across the image plane due to local turbulence strength fluctuations. To synthesize realistic turbulence-degraded imagery, the blur-width $\omega$ must be extended to a spatially varying field constrained by physical turbulence statistics, as elaborated in the following subsection.

\subsubsection{From Blur-Width to Blur-Width Field}\label{subsubsec:blur_field}
To extend the PSF formulation to the full image domain, we define a spatially varying blur-width field $\mathcal{W}(\mathbf{x},\tau)$, where $\mathbf{x}\in\mathbb{R}^2$. This field assigns each spatial location a local blur scale, enabling modeling of spatially non-uniform blur.

For a given exposure-time $\tau$, $\mathcal{W}(\mathbf{x},\tau)$ must be constrained to follow the statistical behavior predicted by optical turbulence theory~\cite{du2015detector}. In this framework, the theoretical mean of the local blur-width is:
\begin{align}
&\overline{\omega}(\tau) =  \frac{2.44\lambda}{D} \times  \notag \\
&\left(1 + 0.268\left(\frac{D}{r_0}\right)^{5/3} + \frac{1.792\left(\frac{D}{r_0}\right)^{5/3}}{1 + \left(6.97\frac{D}{v_w \tau}\right)^{0.607}}\right)^{0.5}.
\label{eq:omega_mean}
\end{align}

Under the same theoretical setting, the corresponding standard deviation is given by:
\begin{align}
\sigma_\omega(\tau) = & 1.70\frac{r_0^{1/3}}{\lambda}\left(D C_n^2 L\right)^{0.5} \times \notag \\
& \left(1 - \frac{1}{1 + \left(4.45\frac{D}{v_w \tau}\right)^{0.5}}\right).
\label{eq:omega_var}
\end{align}

Building on these statistics, we model $\mathcal{W}(\mathbf{x},\tau)$ as a spatially correlated random field with prescribed mean $\overline{\omega}(\tau)$ and standard deviation $\sigma_\omega(\tau)$:
\begin{equation}
\mathcal{W}(\mathbf{x},\tau) 
= \max\!\bigl(\epsilon,\; \overline{\omega}(\tau) 
+ \sigma_\omega(\tau)\,\mathcal{R}(\mathbf{x})\bigr)
\label{eq:W_field}
\end{equation}
where $\mathcal{R}(\mathbf{x})$ is a zero-mean, unit-variance Gaussian random field, and a low-pass filter is applied to introduce spatial correlation. The small lower bound $\epsilon>0$ (e.g., $\epsilon=0.1$\,pixel width) guarantees the physical non-negativity of blur width.

\begin{table*}[ht]
\centering
\caption{Parameter sampling ranges for ET-Turb dataset generation. Notation: $[a,b]$ denotes uniform sampling from continuous interval; $\{v_1, \ldots, v_n\}$ denotes uniform sampling from discrete set. All 12 configurations are sampled with equal probability.}
\small
\setlength{\tabcolsep}{4pt}
\renewcommand{\arraystretch}{0.95}
\begin{tabular}{ccccccc}
\toprule
Distance (m) & Focal length (m) & F-number & $C_n^2$ ($10^{-14}\,\text{m}^{-2/3}$) & Height (m) & Wind speed (m/s) & Exposure-time (ms) \\ 
\midrule
\multirow{2}{*}{[30,100]} & \multirow{2}{*}{[0.1,0.3]} & \{2.8,4\} & [50,300] & \{4,50\} & [1,3] & [0.5,8] \\ 
\cmidrule(lr){3-7}
 & & \{2.8,4,5.6\} & [200,500] & \{100,200\} & [3,5] & [0.5,8] \\ 
\midrule
\multirow{2}{*}{[100,200]} & \multirow{2}{*}{[0.2,0.5]} & \{2.8,4,5.6\} & [5,50] & \{200,400\} & [1,4] & [1,20] \\ 
\cmidrule(lr){3-7}
 & & \{2.8,4,5.6,8\} & [20,100] & \{4,50\} & [2,6] & [0.5,10] \\ 
\midrule
\multirow{2}{*}{[200,400]} & \multirow{2}{*}{[0.3,0.5]} & \{2.8,4,5.6,8\} & [2,30] & \{50,100\} & [2,5] & [1,20] \\ 
\cmidrule(lr){3-7}
 & & \{4,5.6,8,11\} & [10,40] & \{10,50\} & [3,6] & [1,20] \\ 
\midrule
\multirow{2}{*}{[400,600]} & \multirow{2}{*}{[0.4,0.75]} & \{4,5.6,8,11\} & [1,20] & \{50,150\} & [3,5] & [2,40] \\ 
\cmidrule(lr){3-7}
 & & \{5.6,8,11,16\} & [10,30] & \{10,100\} & [4,7] & [1,20] \\ 
\midrule
\multirow{2}{*}{[600,800]} & \multirow{2}{*}{[0.6,0.8]} & \{5.6,8,11,16\} & [1,15] & \{100,300\} & [3,7] & [2,40] \\ 
\cmidrule(lr){3-7}
 & & \{8,11,16,18\} & [2,20] & \{50,200\} & [4,8] & [2,40] \\ 
\midrule
\multirow{2}{*}{[800,1000]} & \multirow{2}{*}{[0.8,1]} & \{8,11,16,18\} & [0.5,10] & \{10,100\} & [5,9] & [2,40] \\ 
\cmidrule(lr){3-7}
 & & \{11,16,18,24\} & [1,20] & \{4,50\} & [6,10] & [1,20] \\ 
\bottomrule
\end{tabular}
\label{tab:parameter_range}
\end{table*}

Equipped with this spatially varying blur-width field, we generate position-dependent PSFs according to Eq.~\eqref{eq:psf_et} and apply them to the tilt image $I_T(\mathbf{x})$ as described in \cite{popkin2010accurate}. The resulting exposure-aware blur operator $\mathcal{B}_\tau$ is implemented as a position-dependent filtering operation:
\begin{equation}
\mathcal{B}_\tau(I_T(\mathbf{x})) = \text{PSF}_{\text{ET}}\!\left(\mathbf{x},\,
\mathcal{W}(\mathbf{x},\tau),\,\tau\right) \ast I_T(\mathbf{x})
\label{eq:Btau_discrete}
\end{equation}
where each kernel $\text{PSF}_{\text{ET}}(\mathbf{x}, \mathcal{W}(\mathbf{x},\tau),\tau)$ is centered at spatial location $\mathbf{x}$ and parameterized by the local blur-width $\mathcal{W}(\mathbf{x},\tau)$ and global exposure-time $\tau$. For simplicity, we still denote this spatially varying filtering operation by $\ast$.

\begin{figure}[htbp]
  \centering
  \includegraphics[width=\columnwidth]{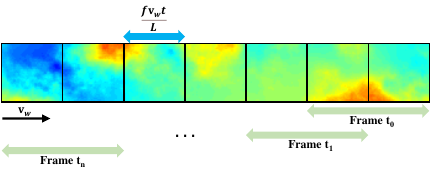} 
  \caption{Temporal evolution of turbulence degradation under Taylor's frozen-flow hypothesis in video synthesis. We construct an extended turbulence degradation field along the wind direction $\mathbf{v}_w$. Temporal correlation between frames is achieved by displacing this field over time using the shift $\frac{f \mathbf{v}_w t}{L}$.} 
  \label{fig:temporal} 
\end{figure}

\subsection{Inter-Frame Correlation}\label{subsec:temporal}
Extending turbulence synthesis from single images to video sequences requires modeling how atmospheric degradation evolves between successive frames. We adopt Taylor's frozen-flow hypothesis~\cite{roggemann1995method}, which treats turbulence as a quasi-stationary refractive index field advected by mean wind, with negligible morphological change. This is valid given the camera exposure-time $\tau \sim (0.5, 40)$ ms, which is much shorter than the turbulence evolution timescale (typically hundreds of milliseconds).

Therefore, the degradation progression across video frames can be modeled as the spatial translation of a frozen degradation field, as illustrated in Fig.~\ref{fig:temporal}. Specifically, turbulence-induced degradation at a given frame time $t$ is related to the initial degradation at $t=0$ through:
\begin{equation}
\mathcal{H}(J_t(\mathbf{x})) =
\mathcal{H}\!\left(J_0\!\left(\mathbf{x} - \frac{f \mathbf{v}_w t}{L}\right)\right)
\label{eq:turb_time}
\end{equation}
where $\mathcal{H}(J_0(\mathbf{x}))$ denotes the turbulence-degraded image at the reference time $t=0$, $\mathbf{v}_w$ is the wind velocity vector, and $\tfrac{f}{L}$ maps physical wind speed to image plane coordinates.

% \vspace{-.2cm}
\section{The ET-Turb Dataset}
\label{sec:dataset}

As mentioned in the previous section, our turbulence synthesis pipeline first applies geometric tilt, followed by an exposure-time-dependent blur process, producing not only fully degraded turbulent images but also intermediate outputs, specifically isolated geometric distortions and isolated blur, enabling more comprehensive training and systematic evaluation of turbulence mitigation algorithms.

We leverage this pipeline to construct \textbf{ET-Turb}, a large-scale synthetic turbulence dataset that explicitly models exposure-time as a continuous variable across diverse imaging conditions. The dataset's sources align with the works we referenced~\cite{zhang2024imaging, zhang2024spatio}, comprising videos sourced from the SVW dataset~\cite{safdarnejad2015sports} and TSR-WGAN~\cite{jin2021neutralizing}, totaling \textbf{5,083 videos} containing \textbf{2,005,835 frames}. These are randomly partitioned into \textbf{3,988 training videos} and \textbf{1,095 test videos} to facilitate robust model development and evaluation. 

\begin{table*}[ht]
\centering
\caption{Quantitative comparison of turbulence mitigation models trained on different synthetic datasets and evaluated on real turbulence data. Models trained on ET-Turb dataset achieve consistently lower NIQE and BRISQUE scores (lower is better), confirming superior realism and generalization capability compared to TMT-Dynamic~\cite{zhang2024imaging} dataset and ATSyn-Dynamic~\cite{zhang2024spatio} dataset.}
\small
\setlength{\tabcolsep}{4pt}
\renewcommand{\arraystretch}{1.3}
\begin{tabular}{lcccccccc}
\hline
\multirow{2}{*}{Training Dataset} & \multicolumn{2}{c}{TSR-WGAN~\cite{jin2021neutralizing}} & \multicolumn{2}{c}{TMT~\cite{zhang2024imaging}} & \multicolumn{2}{c}{DATUM~\cite{zhang2024spatio}} & \multicolumn{2}{c}{MambaTM~\cite{zhang2025learning}} \\ \cline{2-9} 
                                  & NIQE $\downarrow$ & BRISQUE $\downarrow$ & NIQE $\downarrow$ & BRISQUE $\downarrow$ & NIQE $\downarrow$ & BRISQUE $\downarrow$ & NIQE $\downarrow$ & BRISQUE $\downarrow$ \\ \hline
TMT-dynamic~\cite{zhang2024imaging} & 4.231 & 52.502 & 4.361 & 58.581 & 4.219 & 54.921 & 4.217 & 55.062 \\
ATSyn-dynamic~\cite{zhang2024spatio} & 4.224 & 54.462 & 4.483 & 59.707 & 4.308 & 59.126 & 4.247 & 56.876 \\
\textbf{ET-Turb} & \textbf{4.190} & \textbf{50.981} & \textbf{4.221} & \textbf{56.691} & \textbf{4.204} & \textbf{54.070} & \textbf{4.212} & \textbf{55.050} \\ \hline
\end{tabular}
\label{tab:real}
\end{table*}

\begin{figure*}[ht]
    \centering
    % First row of subfigures
    \begin{subfigure}[b]{0.24\textwidth}
        \includegraphics[width=\textwidth]{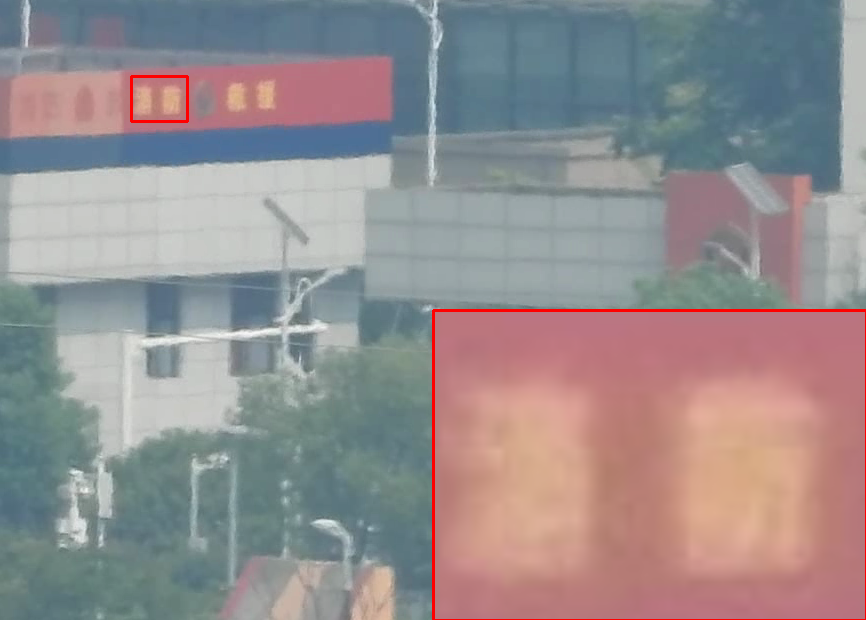}
    \end{subfigure}
    \hfill
    \begin{subfigure}[b]{0.24\textwidth}
        \includegraphics[width=\textwidth]{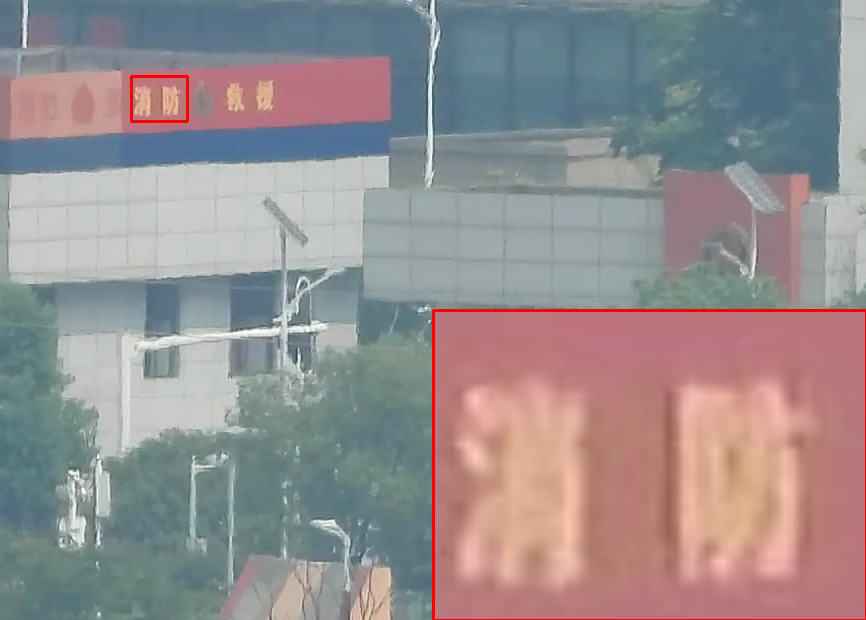}
    \end{subfigure}
    \hfill
    \begin{subfigure}[b]{0.24\textwidth}
        \includegraphics[width=\textwidth]{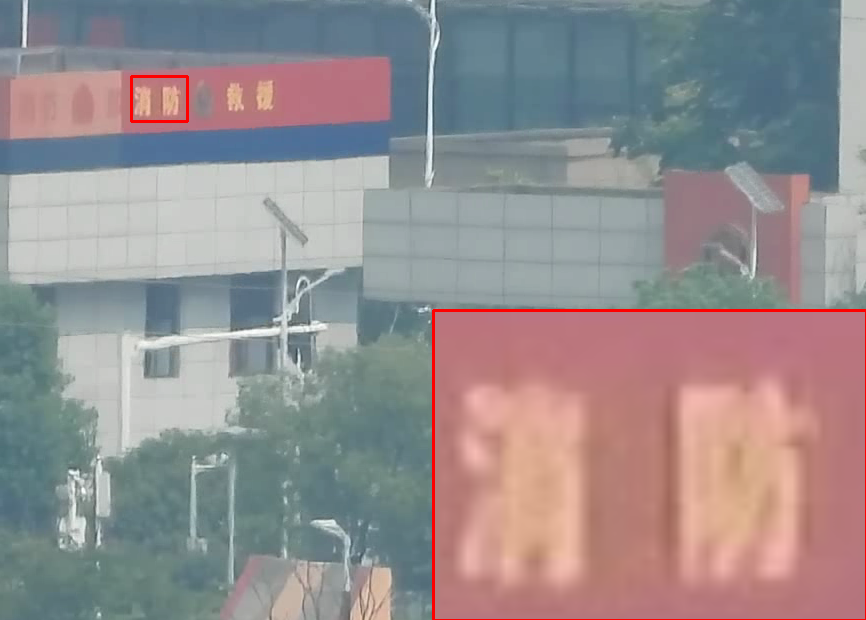}
    \end{subfigure}
    \hfill
    \begin{subfigure}[b]{0.24\textwidth}
        \includegraphics[width=\textwidth]{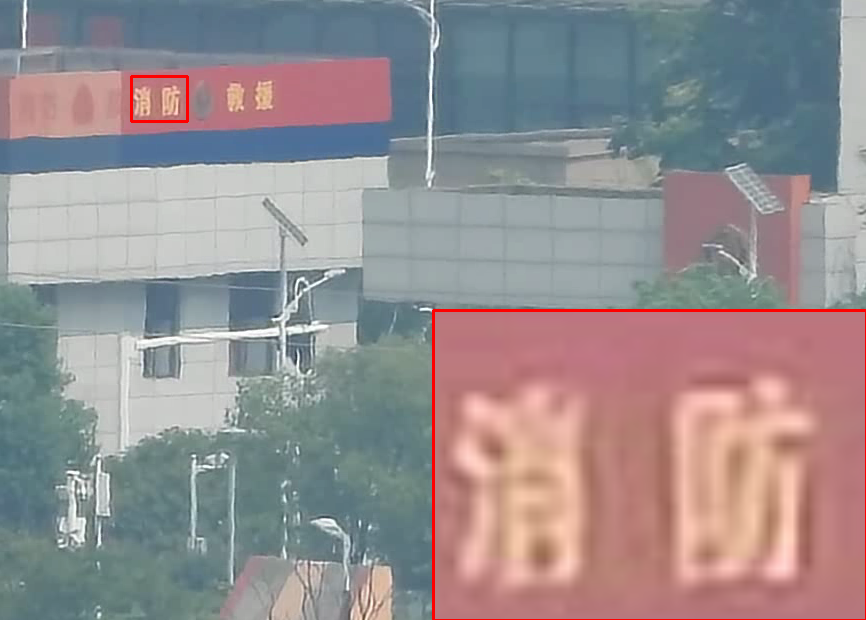}
    \end{subfigure}
    \vspace{0.2cm}
    
    % Second row of subfigures with captions
    \begin{subfigure}[b]{0.24\textwidth}
        \includegraphics[width=\textwidth]{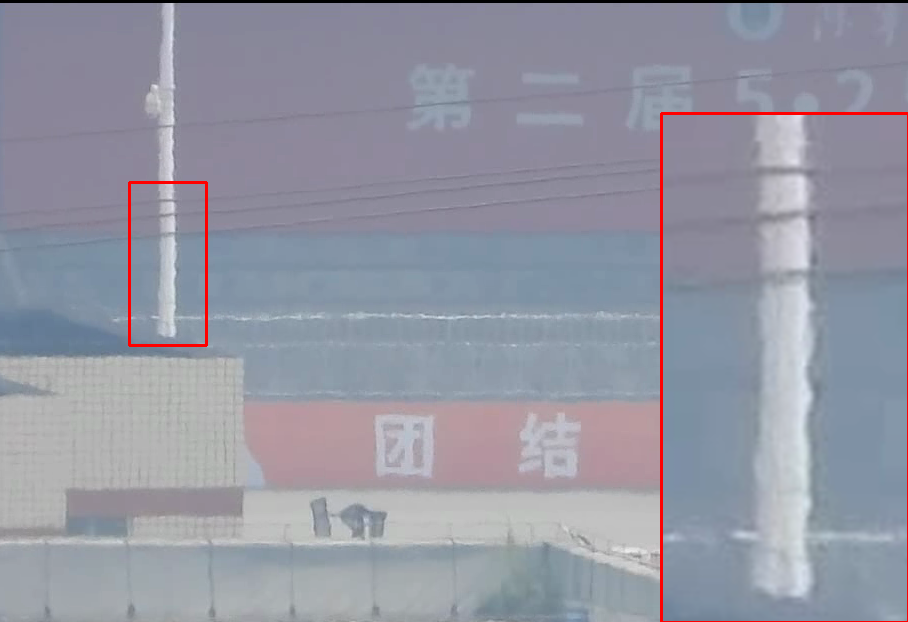}
        \caption{Input}
    \end{subfigure}
    \hfill
    \begin{subfigure}[b]{0.24\textwidth}
        \includegraphics[width=\textwidth]{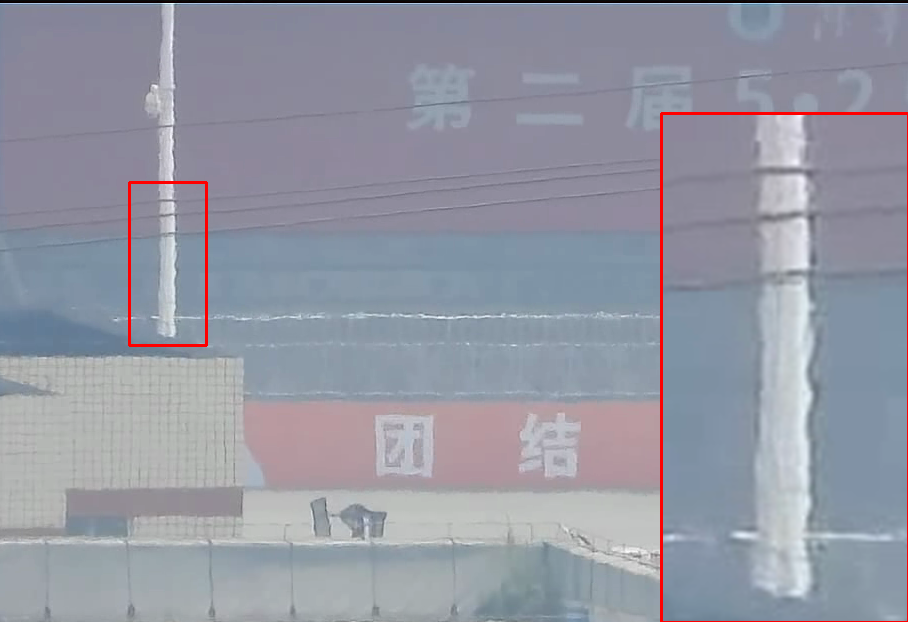}
        \caption{TMT-dynamic~\cite{zhang2024imaging}}
    \end{subfigure}
    \hfill
    \begin{subfigure}[b]{0.24\textwidth}
        \includegraphics[width=\textwidth]{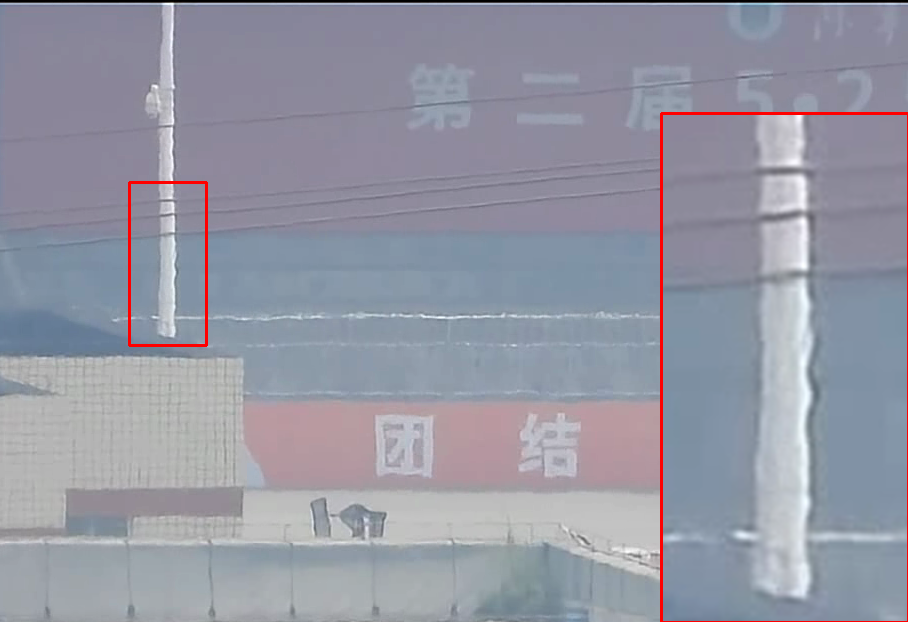}
        \caption{ATSyn-dynamic~\cite{zhang2024spatio}}
    \end{subfigure}
    \hfill
    \begin{subfigure}[b]{0.24\textwidth}
        \includegraphics[width=\textwidth]{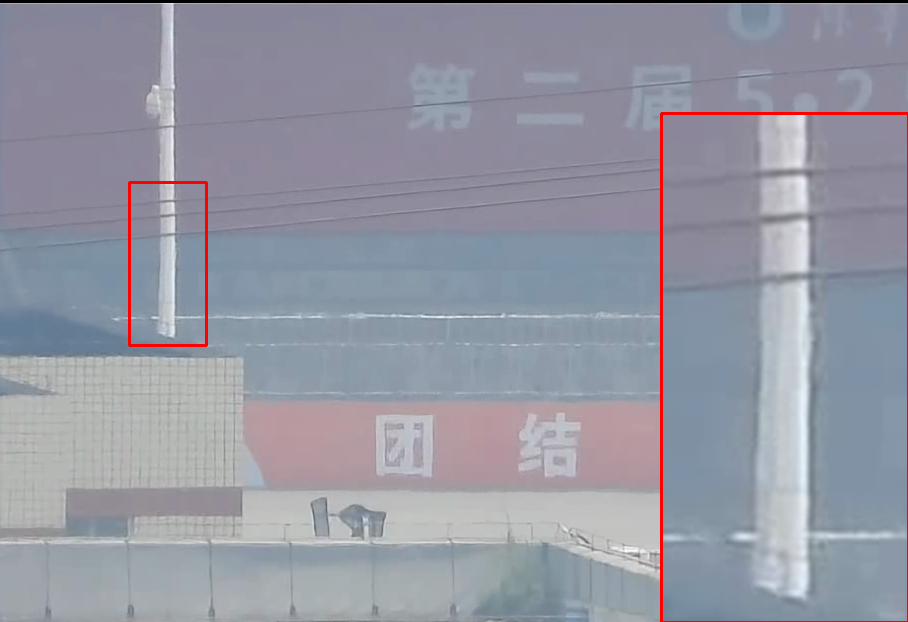}
        \caption{ET-Turb}
    \end{subfigure}
    % \vspace{0.2cm}
    
    \caption{Visual comparison of turbulence mitigation results on real data for MambaTM~\cite{zhang2025learning} models trained on different synthetic datasets. Models trained on ET-Turb dataset produce sharper and more natural restorations with fewer artifacts compared to those trained on TMT-Dynamic~\cite{zhang2024imaging} dataset and ATSyn-Dynamic~\cite{zhang2024spatio} dataset.}
    \label{fig:et_turb_real_comparison}
\end{figure*}
To ensure comprehensive coverage of real-world imaging conditions, we design \textbf{12 turbulence configurations} that systematically span diverse optical and atmospheric regimes, as detailed in Tab.~\ref{tab:parameter_range}. Each configuration is parameterized by seven physical quantities: propagation distance, focal length, aperture F-number, refractive index structure parameter ($C_n^2$), altitude, wind speed, and exposure-time. The configurations uniformly sample the parameter space to ensure balanced coverage, while individual values within each configuration are randomly drawn from physically realistic ranges to maintain both statistical diversity and theoretical consistency. Notably, exposure-time is restricted to \textbf{0.5--40 ms}, consistent with practical imaging systems and avoiding unrealistic extremes rarely encountered in real applications. We provide visual results of the turbulence synthesis pipeline in the supplementary material.

Prior works have established several real-world turbulence datasets under various scenarios. However, these datasets are typically collected using a single capture device with fixed imaging parameters, which limits their capability to comprehensively evaluate model robustness across diverse optical configurations and atmospheric conditions. To address this, we curate a subset of videos from multiple existing datasets~\cite{anantrasirichai2013atmospheric,jin2021neutralizing, xu2024long} to construct the \textbf{ET-Turb-Real} dataset, which consists of \textbf{74 videos} captured by \textbf{three distinct imaging devices} under diverse environmental conditions with varying imaging setups. For more details about ET-Turb-Real, please refer to the supplementary material.

\textbf{ET-Turb} is the synthetic turbulence dataset to explicitly incorporate exposure-time as a continuous modeling parameter throughout the synthesis process. This fundamental advancement substantially narrows the gap between synthetic training data and real-world turbulent observations, enabling significantly improved generalization performance across diverse imaging conditions and establishing a new benchmark for realistic turbulence simulation.

% \vspace{-.2cm}
\section{Experiments}
\label{sec:experiments}
We evaluate four representative state-of-the-art turbulence mitigation algorithms on ET-Turb: TSR-WGAN~\cite{jin2021neutralizing}, TMT~\cite{zhang2024imaging}, DATUM~\cite{zhang2024spatio}, and MambaTM~\cite{zhang2025learning}. All models are trained exclusively on ET-Turb and directly applied to the ET-Turb-Real benchmark without fine-tuning or domain adaptation, enabling a rigorous assessment of synthetic-to-real transferability. Training strictly follows the official configurations of each method, and all models achieve stable convergence. We report comprehensive quantitative and visual results on both the simulated test set and the real-world dataset, with additional experiments and implementation details provided in the supplementary material.

\begin{figure*}[ht]
    \centering
    \begin{subfigure}[b]{0.24\textwidth}
        \includegraphics[width=\textwidth]{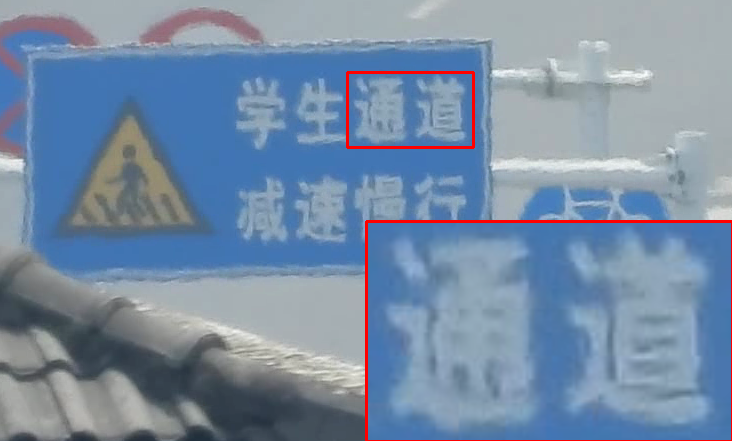}
        \caption{Input}
    \end{subfigure}
    \hfill
    \begin{subfigure}[b]{0.24\textwidth}
        \includegraphics[width=\textwidth]{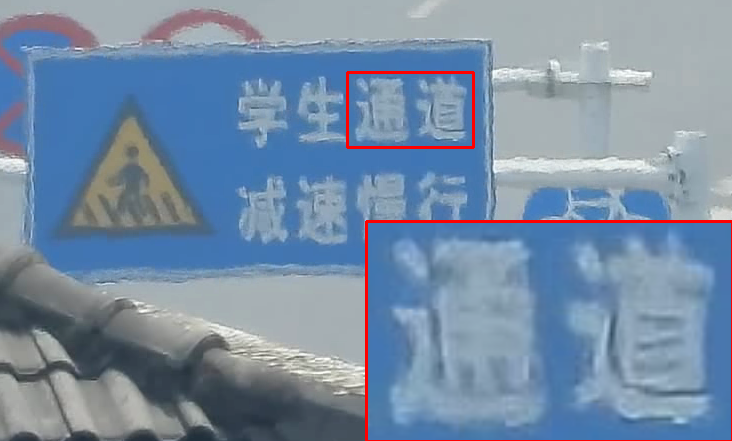}
        \caption{$\tau = 1$ms}
    \end{subfigure}
    \hfill
    \begin{subfigure}[b]{0.24\textwidth}
        \includegraphics[width=\textwidth]{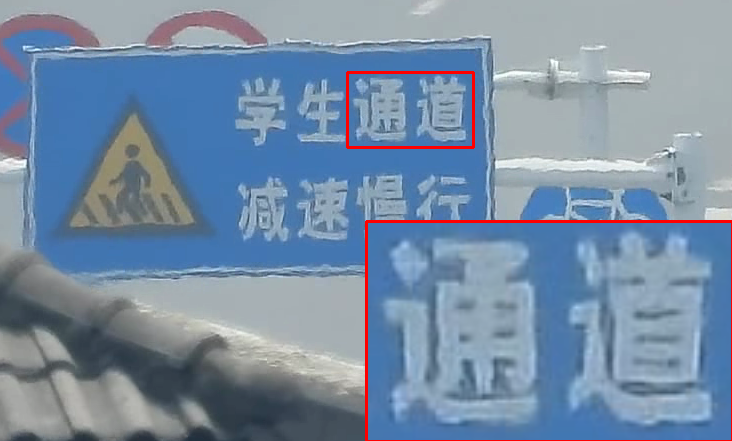}
        \caption{MTF$_{\text{SE/LE}}$}
    \end{subfigure}
    \hfill
    \begin{subfigure}[b]{0.24\textwidth}
        \includegraphics[width=\textwidth]{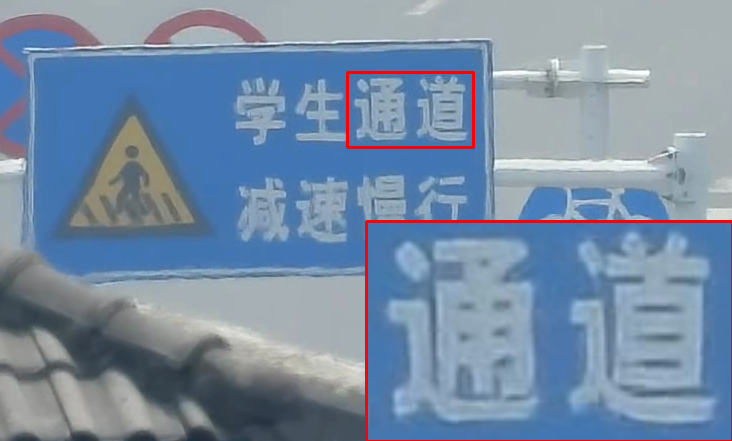}
        \caption{MTF$_{\text{ET}}$}
    \end{subfigure}
    
    \caption{Visual ablation study on exposure modeling strategies for real turbulence restoration. The continuous ET-MTF formulation produces more natural textures and stable structures compared to fixed or binary exposure approaches.}
    \label{fig:ablation}
\end{figure*}

To validate the advantages of ET-Turb over existing synthetic turbulence datasets, we conduct comparative experiments against two widely recognized benchmarks: TMT-dynamic~\cite{zhang2024imaging} and ATSyn-dynamic~\cite{zhang2024spatio}, using identical training protocols and experimental setups. These models are then evaluated on our real-world turbulence dataset for further comparative analysis. Due to the lack of ground truth references in real turbulence data, we utilize no-reference image quality metrics, including NIQE~\cite{mittal2012making} and BRISQUE~\cite{mittal2011blind}, for quantitative evaluation.

\subsection{Comparison with Existing Synthetic Datasets}
\label{subsec:dataset_comparison}

To ensure fair comparisons and eliminate potential training artifacts, all models are trained using their official configurations and evaluated on the datasets involved in the experiments, thereby obtaining the corresponding model files. The models are then applied in a zero-shot manner to the ET-Turb-Real dataset for cross-domain evaluation, providing an objective comparison of performance across domains. Quantitative results are summarized in Tab.~\ref{tab:real}, with qualitative comparisons provided in Fig.~\ref{fig:et_turb_real_comparison}.

The high physical fidelity of the ET-Turb dataset enables models trained on it to effectively generalize to real turbulent imagery, producing naturally colored and structurally coherent de-turbulence results. In contrast, models trained on alternative datasets exhibit limited generalization capability, often introducing artifacts and residual blur in restored images, such as distorted text on buildings or distant power poles. Quantitative results consistently demonstrate that, across all evaluated models, training on ET-Turb yields superior performance compared to other datasets, highlighting its enhanced practicality and generalization.

\subsection{Ablation Study}
\label{subsec:ablation_study}

To validate the effectiveness of the proposed ET-MTF, we generate two ablation datasets using identical clean images from ET-Turb but with alternative exposure modeling strategies: (a) a fixed 1~ms exposure for all samples, and (b) a binary short/long-exposure MTF (MTF$_{\text{SE/LE}}$). All other parameters remain unchanged. We train MambaTM~\cite{zhang2025learning} on each variant and evaluate on ET-Turb-Real, with results shown in Fig.~\ref{fig:ablation} and Tab.~\ref{tab:ablation}.

\begin{table}[t]
\centering
\caption{Ablation study on exposure modeling strategies within the ET-Turb synthesis pipeline. Comparing fixed ($\tau = 1$ms), binary (MTF$_{\text{SE/LE}}$), and continuous (MTF$_{\text{ET}}$) exposure formulations demonstrates that continuous modeling yields the best perceptual quality on MambaTM~\cite{zhang2025learning}.}
\begin{tabular}{lcc}
\hline
Exposure Strategy & NIQE $\downarrow$ & BRISQUE $\downarrow$ \\ 
\hline
ET-Turb ($\tau = 1$ms) & 4.355 & 55.457 \\
ET-Turb (MTF$_{\text{SE/LE}}$) & 4.297 & 55.123 \\
ET-Turb (MTF$_{\text{ET}}$) & \textbf{4.212} & \textbf{55.050} \\ 
\hline
\end{tabular}
\label{tab:ablation}
\end{table}

The results highlight distinct performance characteristics across the different exposure modeling strategies. The fixed 1~ms exposure model struggles significantly to restore strong blur due to the insufficient variation in training data across different exposure conditions. The MTF$_{\text{SE/LE}}$ model improves restoration but still exhibits residual blur, indicating its limited coverage of intermediate exposure regimes. In contrast, our ET-Turb with continuous ET-MTF achieves the most natural and visually coherent restoration, with superior blur removal, clearly demonstrating the critical importance of modeling exposure-time as a continuous variable for achieving realistic turbulence synthesis.

% \vspace{-.3cm}
\section{Conclusion}
This paper presents a physics-based approach to atmospheric turbulence synthesis that continuously models turbulence-induced blur as a function of exposure-time. By deriving an Exposure-Time-dependent MTF (ET-MTF), we bridge the gap between the classical short and long-exposure regimes through smooth interpolation, overcoming the limitations of discrete exposure modeling in prior work. We further develop a tilt-invariant PSF combined with a spatially varying blur-width field, establishing a unified and physically-grounded representation of turbulence blur. Building on this pipeline, we construct ET-Turb, a large-scale synthetic turbulence dataset to incorporate exposure-time as a continuous variable in the synthesis process. Experiments demonstrate that models trained on ET-Turb achieve superior generalization on real turbulence data, outperforming those trained on existing datasets in both visual quality and no-reference metrics, validating the enhanced realism and practical utility of our exposure-aware synthesis framework.

Our work provides a more reliable foundation for training data generation in turbulence simulation and image restoration. Moreover, the continuous exposure modeling framework establishes a pathway toward future research in dynamic exposure-conditioned turbulence synthesis.

% \section{Acknowledgments}
% This work is supported in part by the National Natural Science Foundation of China under grants 62272229, 62472224 and U2441285, the Natural Science Foundation of Jiangsu Province under grants BK20222012, and Shenzhen Science and Technology Program JCYJ20230807142001004 and Shenzhen Longhua Science and Technology Innovation Special Funding Project (Industrial Sci-Tech Innovation Center of Low-Altitude Intelligent Networking).

{
    \small
    \bibliographystyle{ieeenat_fullname}
    \bibliography{main}
}
\newpage
\appendix
\begin{figure*}[!t]
    \centering
    \begin{subfigure}[b]{0.32\textwidth}
        \includegraphics[width=\textwidth]{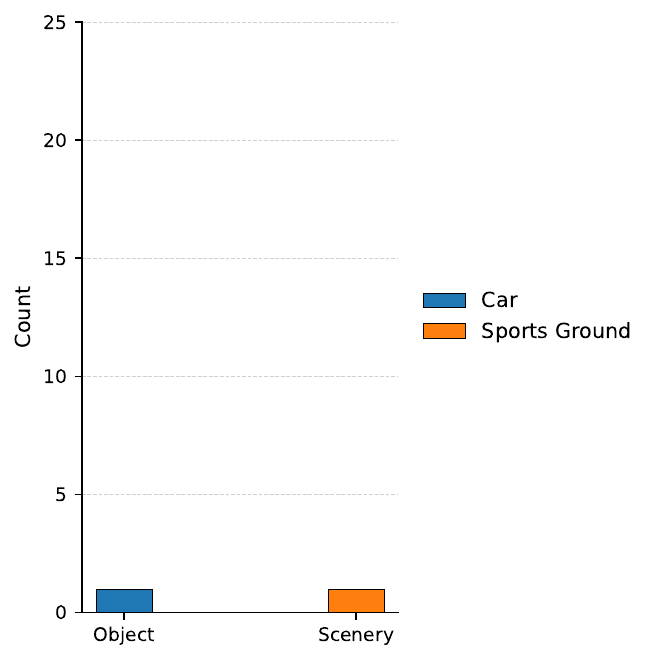}
        \caption{Captured by Canon EOS-1D Mark IV~\cite{anantrasirichai2013atmospheric}}
    \end{subfigure}
    \hfill
    \begin{subfigure}[b]{0.32\textwidth}
        \includegraphics[width=\textwidth]{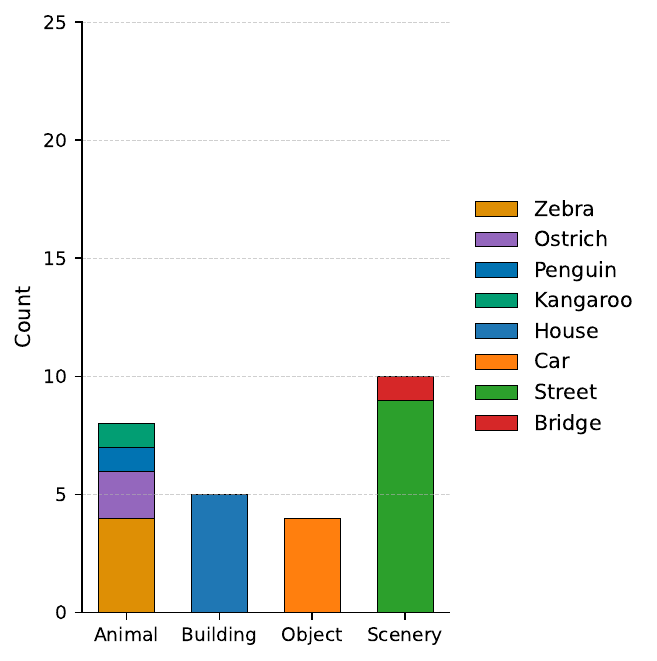}
        \caption{Captured by Nikon D750~\cite{jin2021neutralizing}}
    \end{subfigure}
    \hfill
    \begin{subfigure}[b]{0.32\textwidth}
        \includegraphics[width=\textwidth]{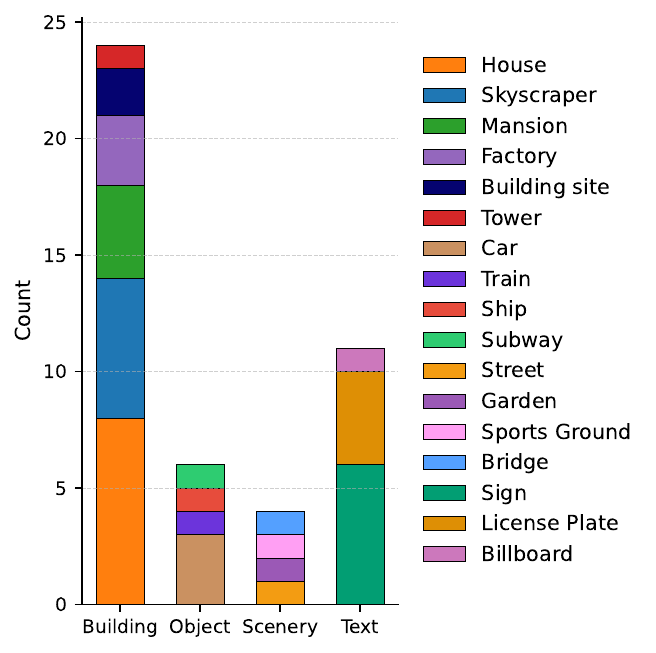}
        \caption{Captured by Nikon Coolpix P1000~\cite{xu2024long}}
    \end{subfigure}
    
    \caption{ET-Turb-Real dataset composition: multi-camera systems and diverse scene distribution. Real-world turbulence data is collected across three different imaging devices, providing comprehensive coverage of diverse environmental and anthropogenic scenes such as urban landscapes, entertainment venues, industrial infrastructure, traffic scenarios, and commercial establishments.}
    \label{fig:ET-real}
\end{figure*}

\section{Details of the Proposed ET-Turb Dataset}
\subsection{Visualization of the Turbulence Degradation Process}
Our data synthesis method decomposes atmospheric turbulence degradation into two components: tilt and blur~\cite{shimizu2008super,leonard2012simulation,chan2022tilt,charnotskii2022warp,lau2019restoration}. This decomposition enables quantitative analysis and provides diverse training data with controllable degradation parameters for mitigation algorithms. In this section, we visualize the synthesis pipeline and demonstrate how configuration parameters affect image quality.

\textbf{Tilt Component:} Fig.~\ref{fig:col-tilt} illustrates tilt effects from low-order wavefront aberrations, causing statistically correlated pixel displacements that vary non-uniformly across the image. The color-coded displacement field indicates the magnitude of pixel displacements.

\textbf{Blur Component:} Fig.~\ref{fig:col-blur} shows the exposure-dependent blur applied to the tilt image. The blur intensity depends on exposure-time $\tau$, with longer exposures accumulating more turbulence. By integrating our ET-MTF formulation with spatially varying blur kernels, we simulate realistic, temporally and spatially varying turbulence blur.

\textbf{Complete Turbulence Synthesis:} Fig.~\ref{fig:col-turb} presents the final turbulence-degraded images from the combined tilt and blur components, providing high-quality training data for deep learning-based mitigation algorithms.

The synthesis pipeline is demonstrated across six distinct parameter configurations, as shown in Tab.~\ref{supp:parameter_range}, highlighting the flexibility of our approach in generating a wide range of turbulence conditions. All parameters are selected randomly, and the complete synthesis process is presented, further demonstrating the broad applicability of our method in simulating real-world scenarios.

\subsection{Generation of Turbulent Images under Varying Intensities and Exposure-Time}

\begin{figure*}[p]
    \centering
    % ===== Row 1: C_n^2 labels =====
    \begin{minipage}[b]{0.56\textwidth}
        \centering
        \makebox[0.48\textwidth]{\textbf{$\mathsf{C_n^2 = 5\times10^{-13}\,\mathrm{m}^{-2/3}}$}}
        \hfill
        \makebox[0.48\textwidth]{\textbf{$\mathsf{C_n^2 = 1\times10^{-12}\,\mathrm{m}^{-2/3}}$}}
    \end{minipage}
    % \vspace{6pt}
    
    % ===== Row 2: τ = 1 ms =====
    \begin{subfigure}[b]{0.56\textwidth}
        \centering
        \includegraphics[width=0.48\textwidth]{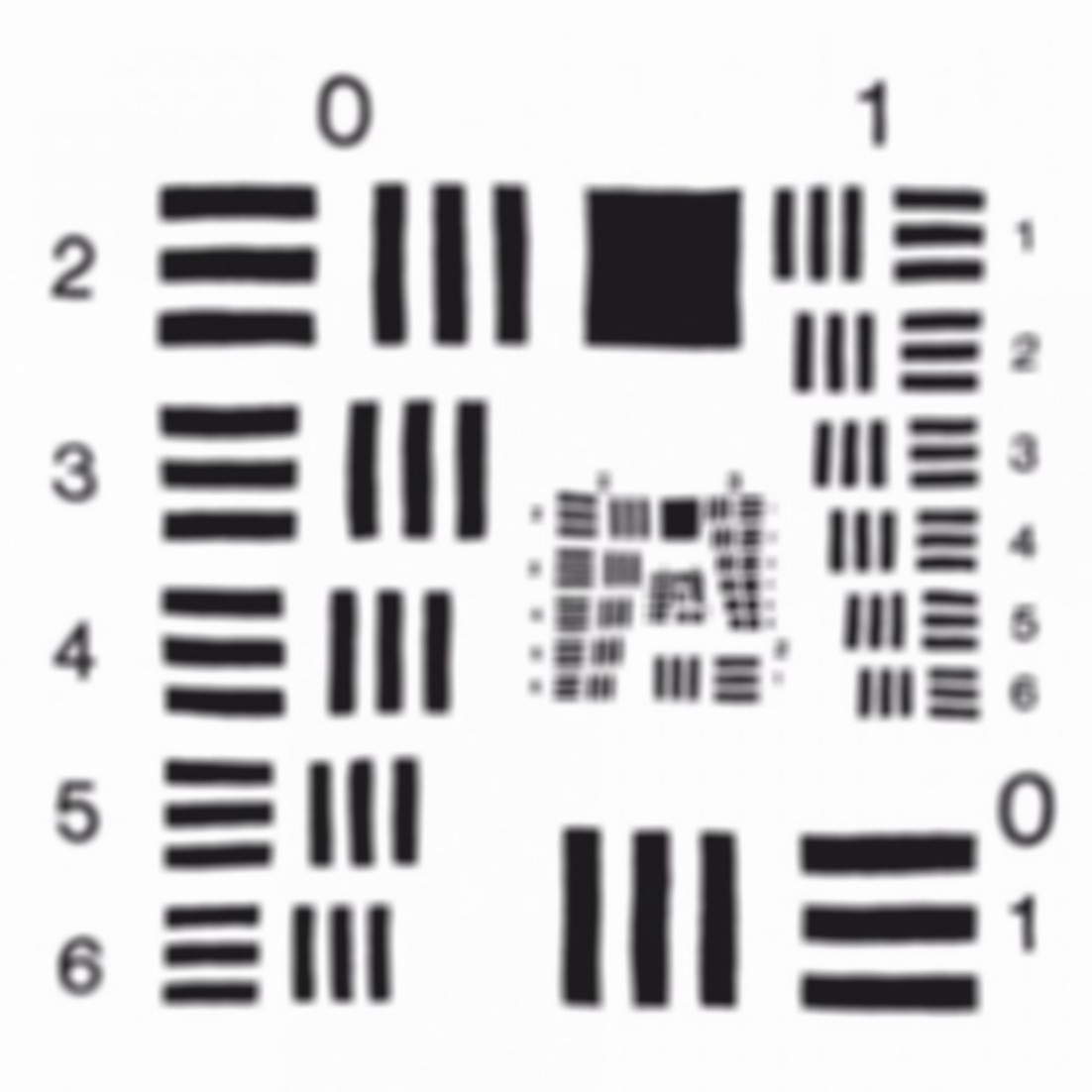}
        \hfill
        \includegraphics[width=0.48\textwidth]{fig/fig_pic_8/2.jpg}
        \caption{$\tau$ = 1\,ms}
    \end{subfigure}
    % \vspace{2pt}
    
    % ===== Row 3: τ = 4 ms =====
    \begin{subfigure}[b]{0.56\textwidth}
        \centering
        \includegraphics[width=0.48\textwidth]{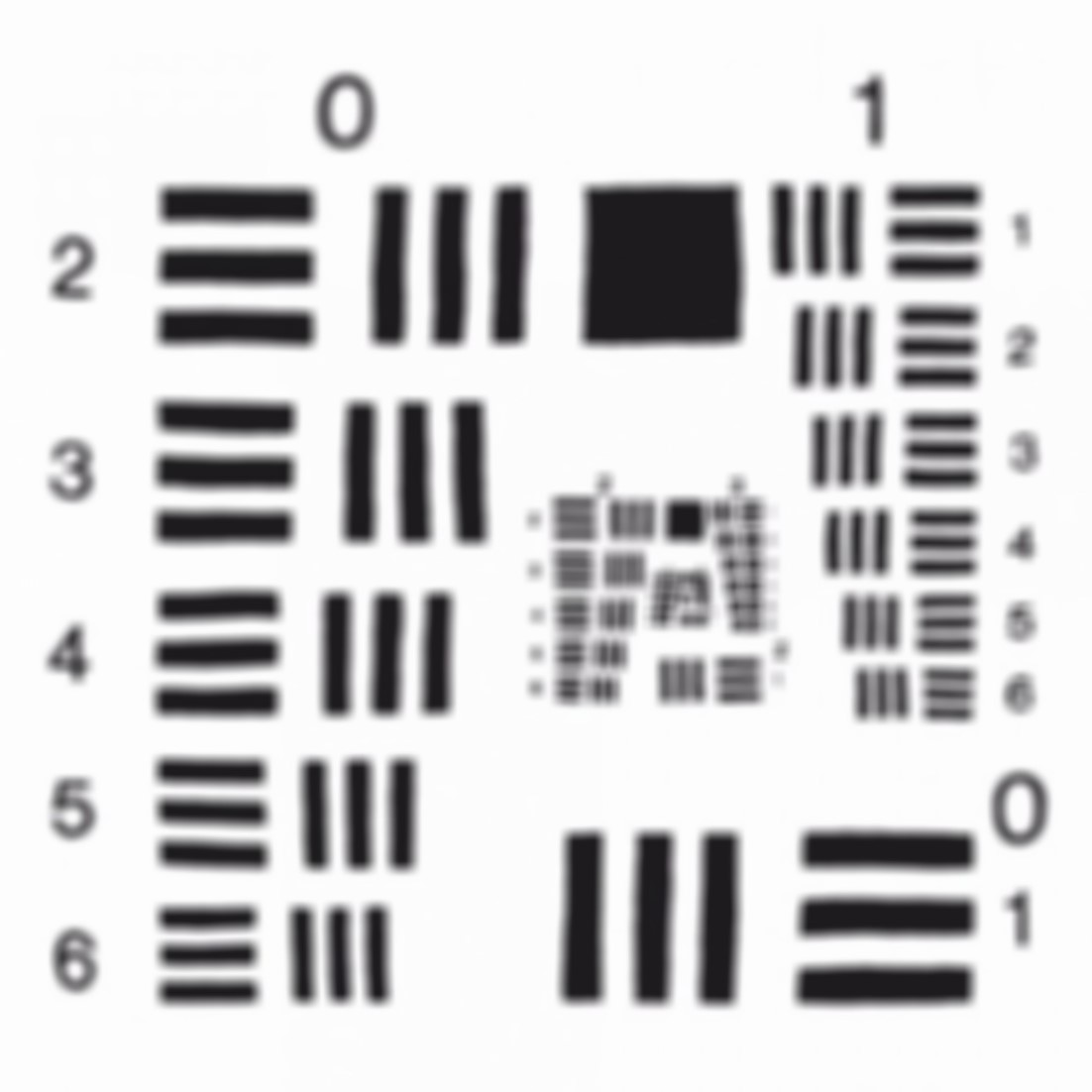}
        \hfill
        \includegraphics[width=0.48\textwidth]{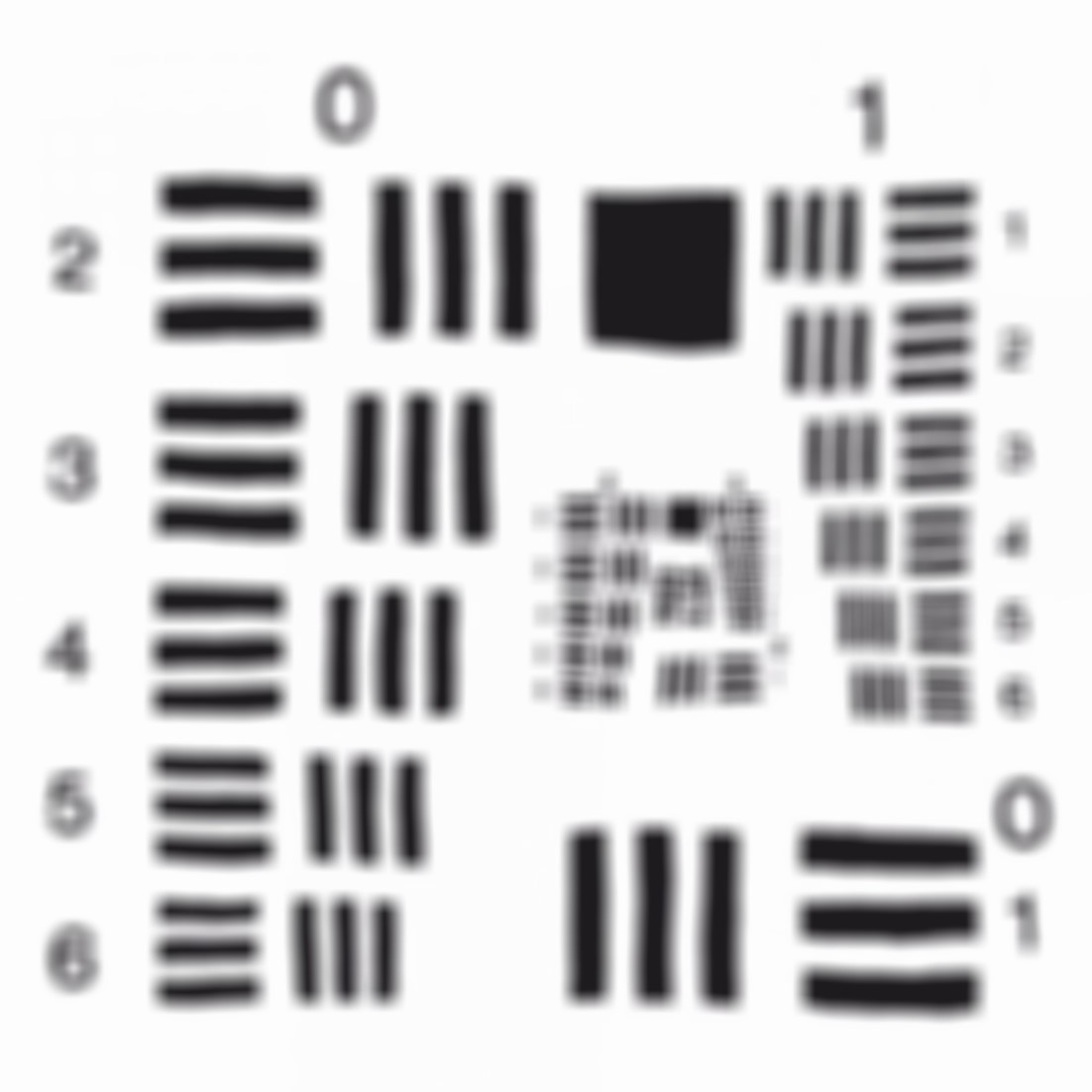}
        \caption{$\tau$ = 4\,ms}
    \end{subfigure}
    % \vspace{2pt}
    
    % ===== Row 4: τ = 10 ms =====
    \begin{subfigure}[b]{0.56\textwidth}
        \centering
        \includegraphics[width=0.48\textwidth]{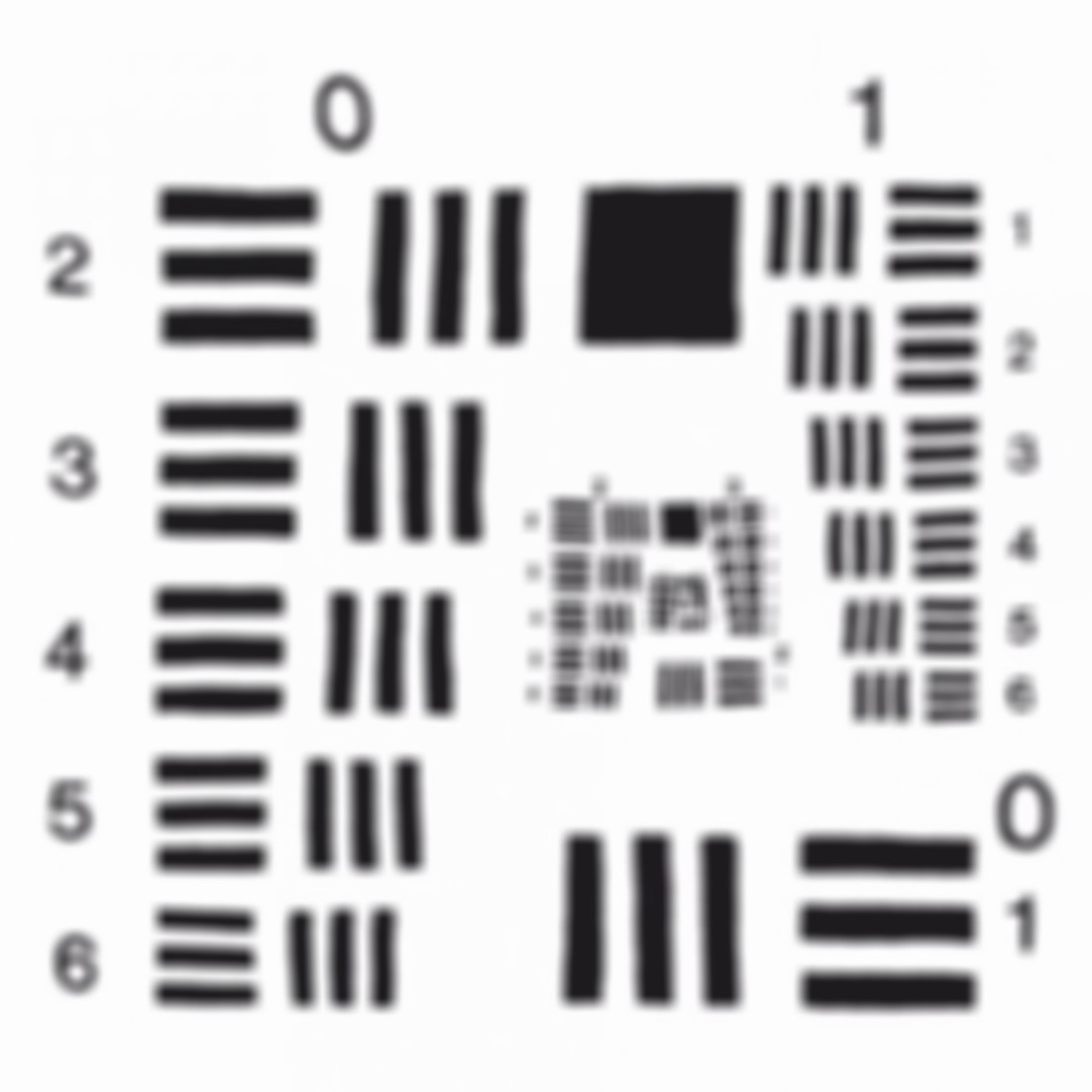}
        \hfill
        \includegraphics[width=0.48\textwidth]{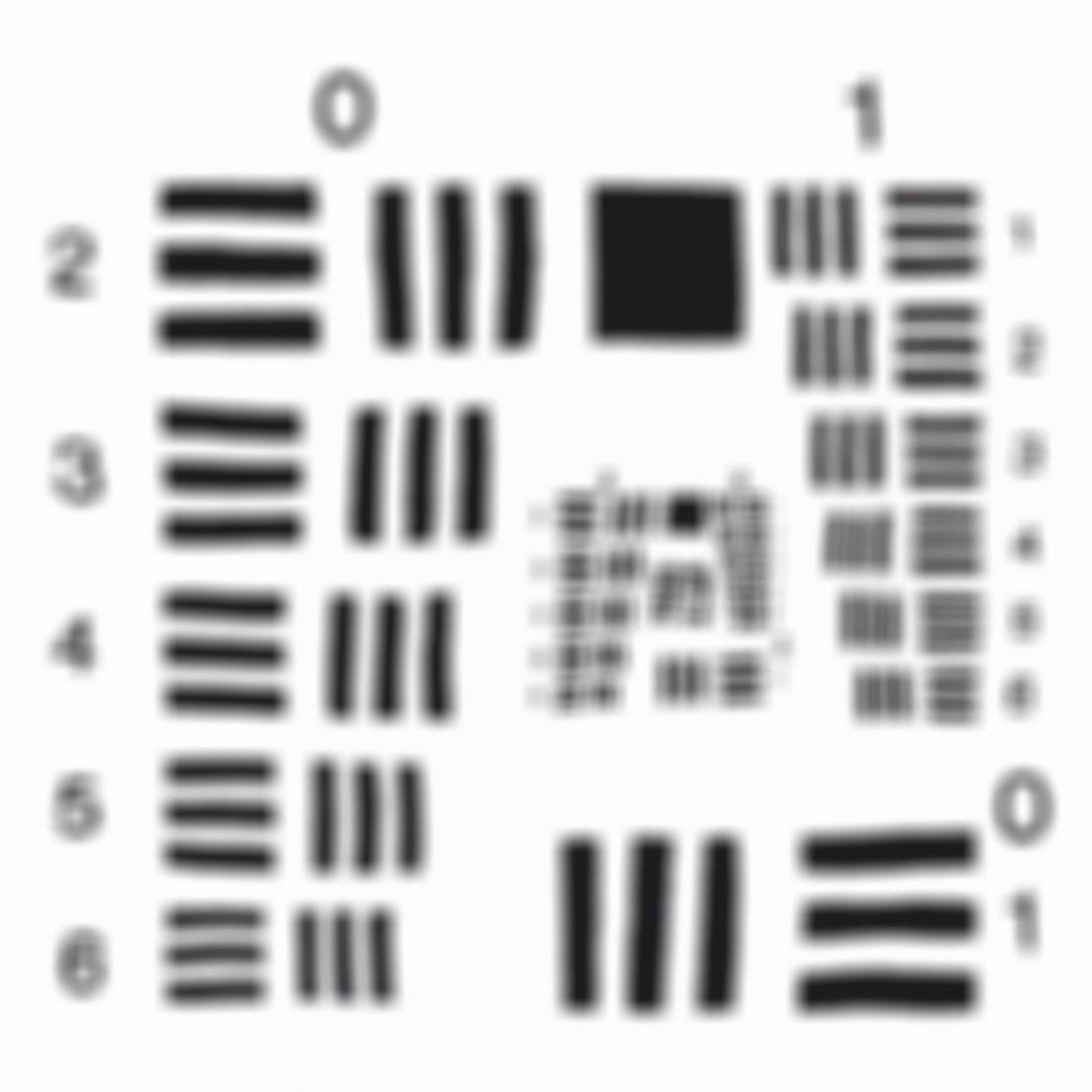}
        \caption{$\tau$ = 10\,ms}
    \end{subfigure}
    % \vspace{pt}
    
    % ===== Row 5: τ = 40 ms =====
    \begin{subfigure}[b]{0.56\textwidth}
        \centering
        \includegraphics[width=0.48\textwidth]{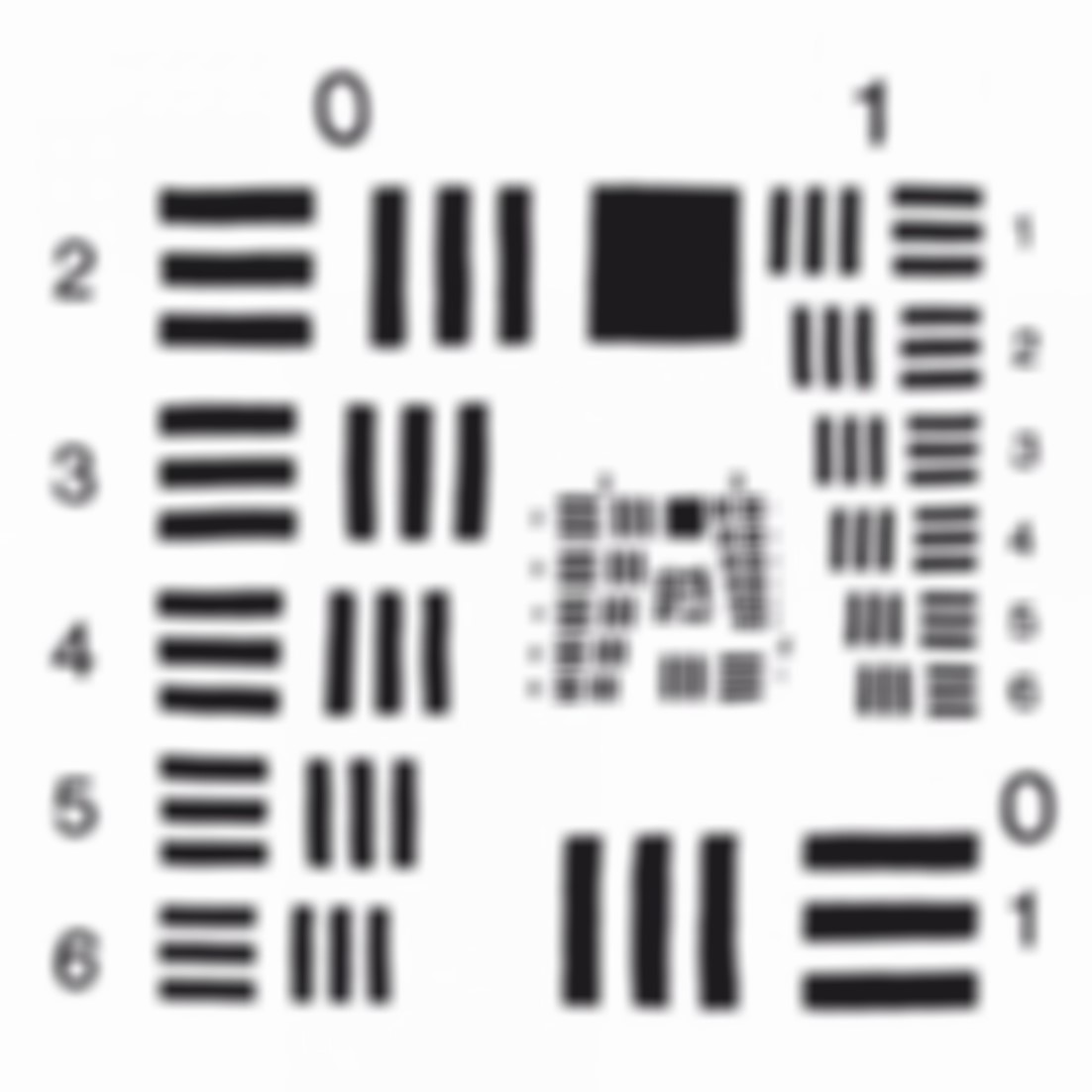}
        \hfill
        \includegraphics[width=0.48\textwidth]{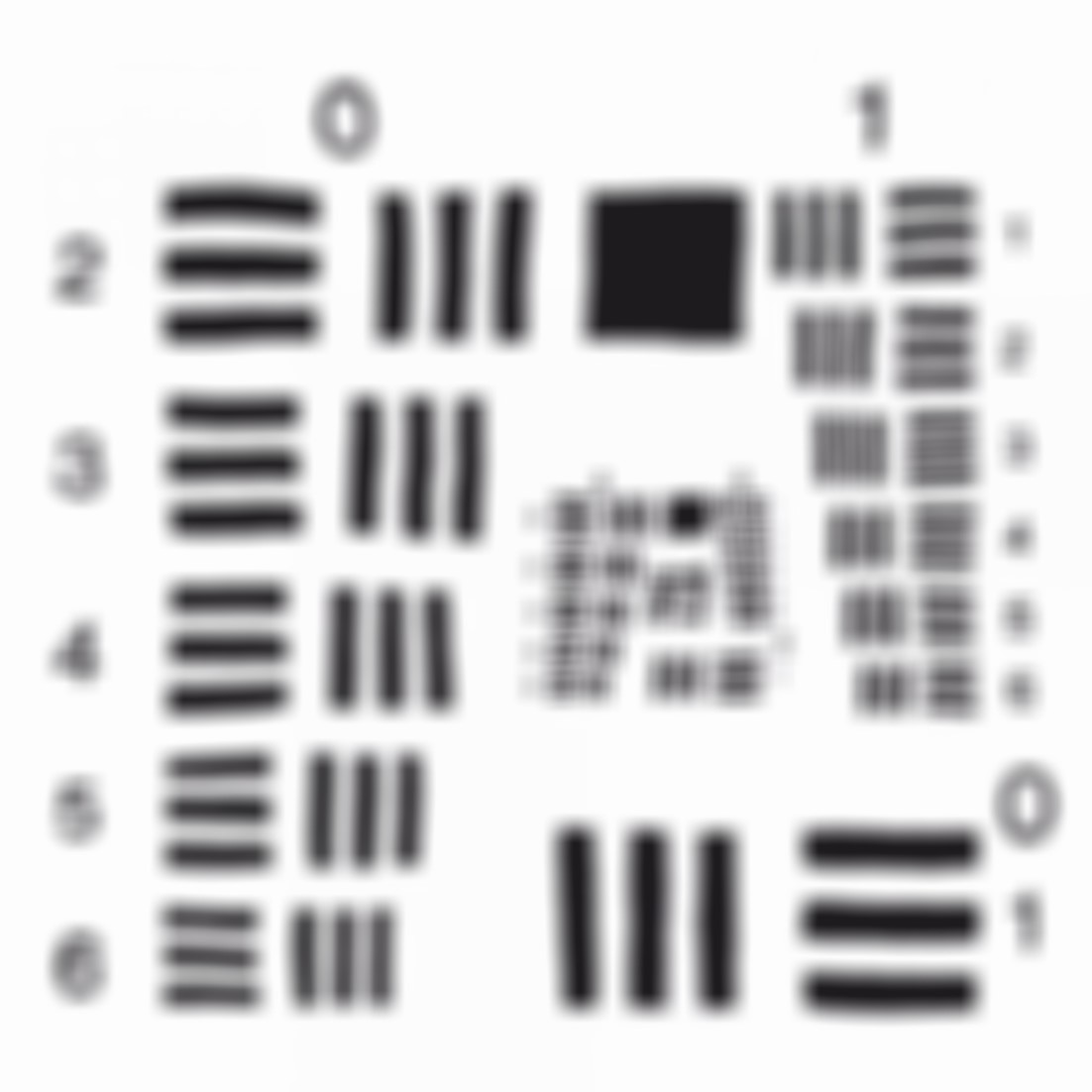}
        \caption{$\tau$ = 40\,ms}
    \end{subfigure}
    % \vspace{2pt}
    
    \caption{Generation of turbulent images under varying intensities and exposure-time using our data synthesis method. As the exposure-time and turbulence intensity increase, the degree of blur becomes more pronounced. The associated parameter set is distance = 500\,m, focal length = 300\,mm, F-number = 8, height = 50\,m, and wind speed = 5\,m/s.}
    \label{supp:fig_exposure_intensity}
\end{figure*}

To further demonstrate the flexibility and physical fidelity of our data synthesis framework, Fig.~\ref{supp:fig_exposure_intensity} shows turbulence-degraded images of a resolution target under varying turbulence intensities and exposure-time. As both exposure-time and turbulence strength increase, the degree of blur becomes progressively more pronounced, reflecting the cumulative effect of turbulence states. For this visualization, we fix the distance to 500~m, the focal length to 300~mm, the F-number to 8, the height to 50~m, and the wind speed to 5~m/s, corresponding to a practically relevant ground-to-ground imaging scenario commonly encountered in long-range surveillance and outdoor observation. These results illustrate that our pipeline reproduces the expected interplay between exposure duration, turbulence strength, and image degradation across a broad range of operating conditions, providing qualitative evidence for the physical plausibility and robustness of our turbulence synthesis approach.

\subsection{Analysis of ET-Turb-Real}

ET-Turb-Real is a real-world turbulence dataset, compiled from publicly available atmospheric turbulence datasets~\cite{anantrasirichai2013atmospheric,jin2021neutralizing,xu2024long}. This dataset is used to evaluate the cross-domain generalization of models trained on synthetic data, providing diverse real-world turbulence-degraded video sequences captured under various environmental and optical conditions.

The dataset spans a wide range of scene categories recorded in different settings, including urban (streets, buildings), recreational (sports fields), industrial (factories), transportation (vehicles), and commercial (billboards). Data acquisition was performed using professional imaging systems such as the Canon EOS-1D Mark IV, Nikon D750, and Nikon Coolpix P1000. These datasets were collected with consistent camera parameters, and we consolidated them to maximize scene diversity while maintaining uniform imaging conditions, ensuring a comprehensive representation of real-world turbulence.

Fig.~\ref{fig:ET-real} shows the dataset composition with stacked bar charts, illustrating the distribution of camera models and scene categories. These charts visualize the proportion of different scene types captured by various imaging devices, ensuring the dataset covers turbulence effects across different scales and environments, essential for evaluating the robustness and generalization of turbulence mitigation models.

The combination of diverse scene types and imaging conditions enhances the dataset's representativeness of real-world turbulence scenarios, thereby establishing a realistic and diverse evaluation set for developing and evaluating atmospheric turbulence mitigation algorithms.

\section{Supplementary Experiments}
\label{sec:supplementary_experiments}

\subsection{Real-World Turbulence Mitigation Performance}
\begin{figure*}[p]
    \centering
    % ===== Row 1: Input =====
    \begin{subfigure}[b]{0.73\textwidth}
        \centering
        \includegraphics[height=3.7cm]{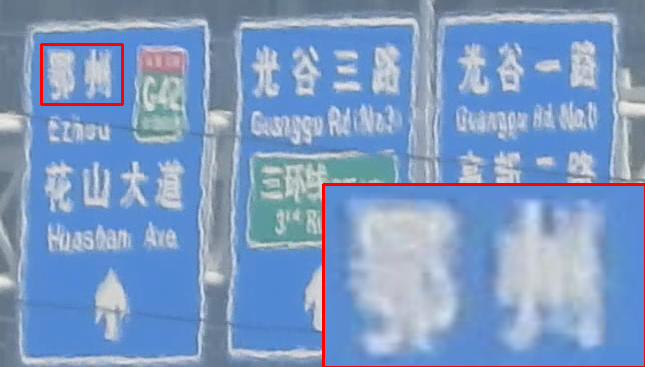}
        \hfill
        \includegraphics[height=3.7cm]{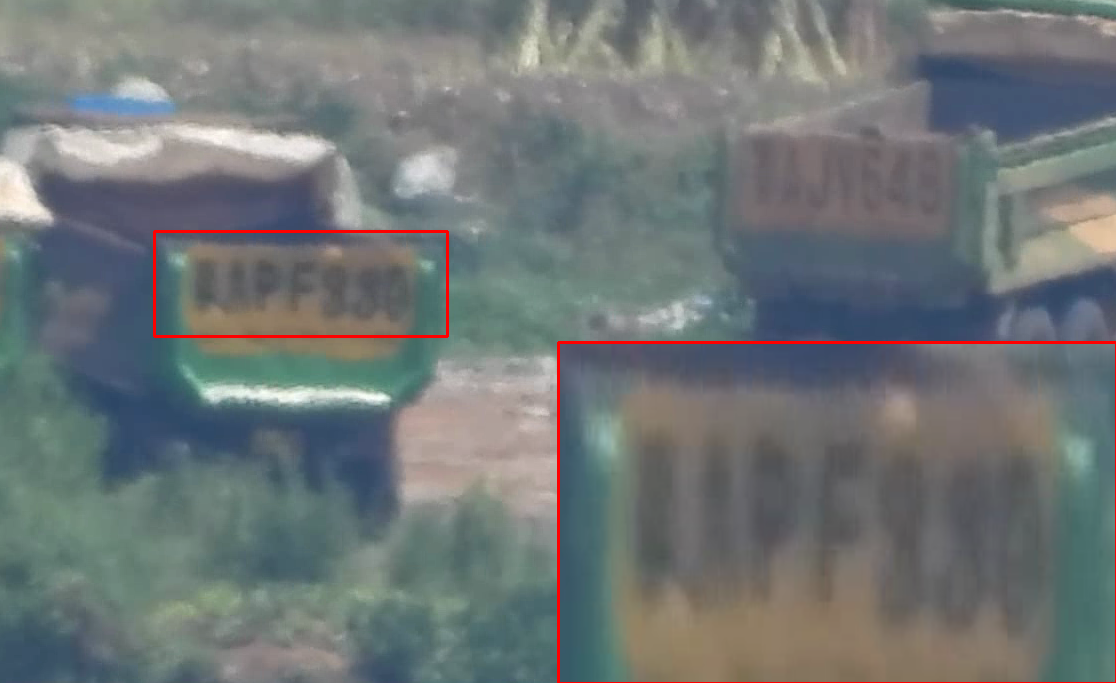}
        \caption{Input}
    \end{subfigure}
    % \vspace{2pt}
    
    % ===== Row 2: TSR-WGAN =====
    \begin{subfigure}[b]{0.73\textwidth}
        \centering
        \includegraphics[height=3.7cm]{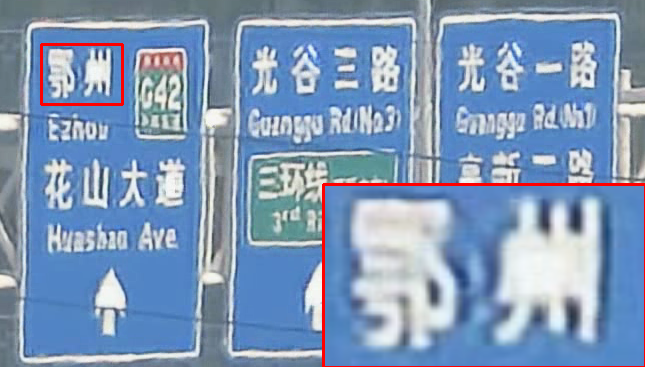}
        \hfill
        \includegraphics[height=3.7cm]{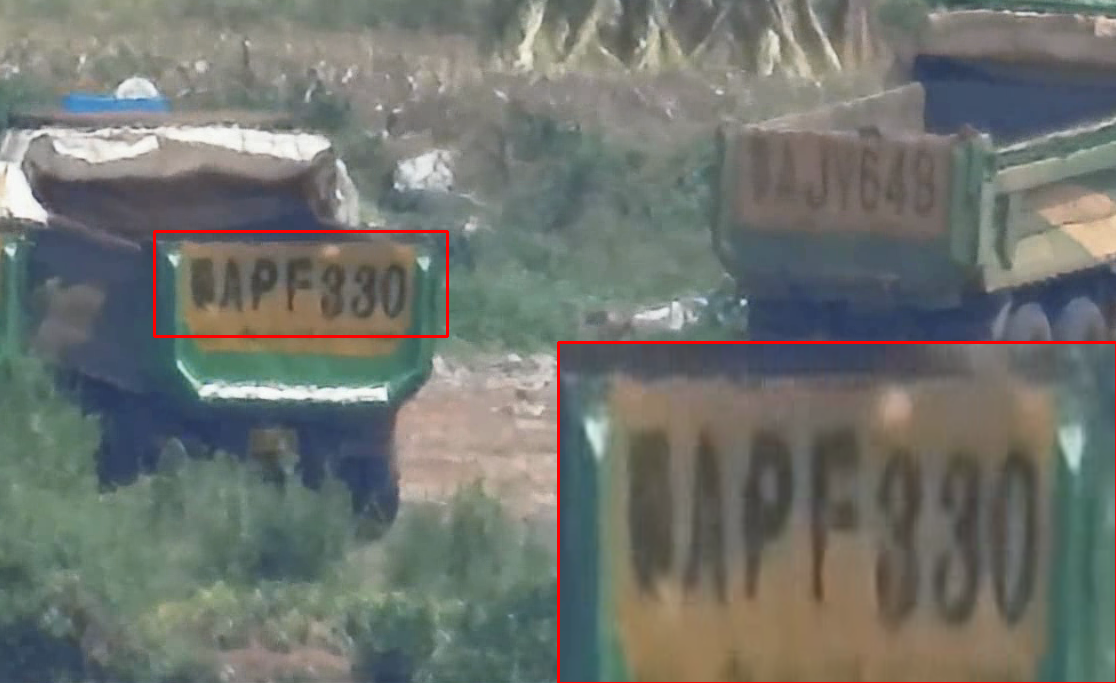}
        \caption{TSR-WGAN~\cite{jin2021neutralizing}}
    \end{subfigure}
    % \vspace{2pt}
    
    % ===== Row 3: TMT =====
    \begin{subfigure}[b]{0.73\textwidth}
        \centering
        \includegraphics[height=3.7cm]{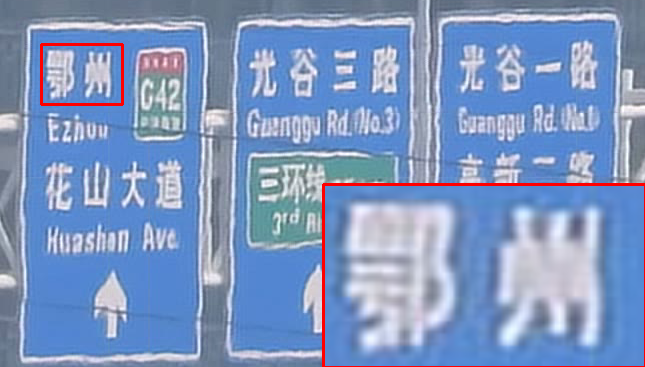}
        \hfill
        \includegraphics[height=3.7cm]{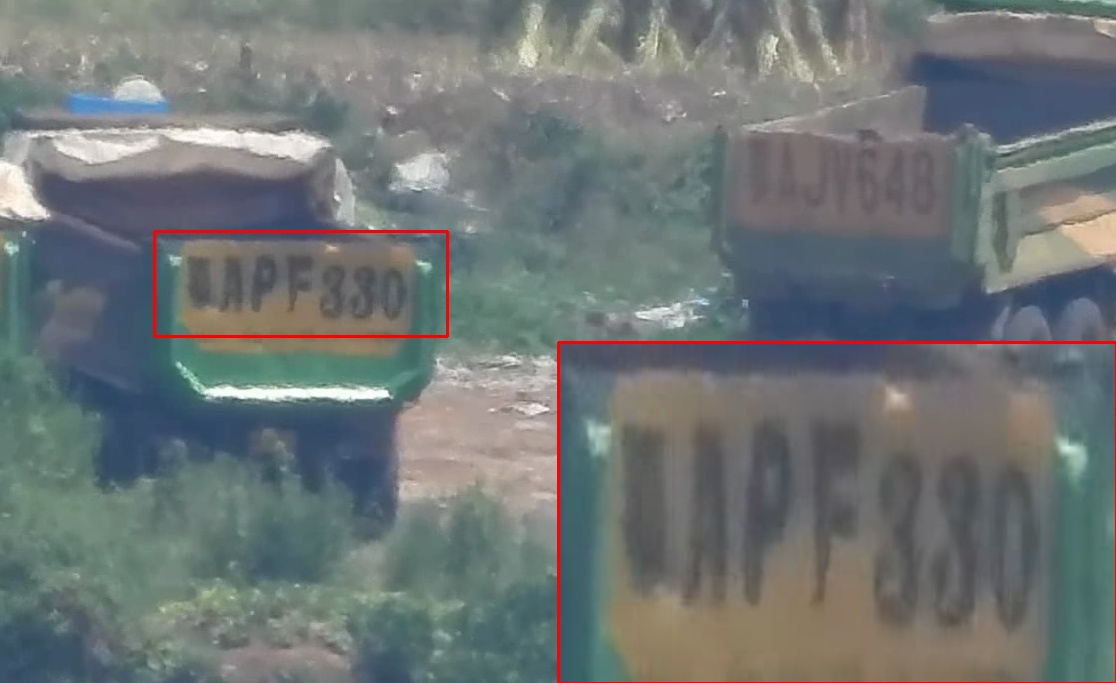}
        \caption{TMT~\cite{zhang2024imaging}}
    \end{subfigure}
    % \vspace{2pt}
    
    % ===== Row 4: DATUM =====
    \begin{subfigure}[b]{0.73\textwidth}
        \centering
        \includegraphics[height=3.7cm]{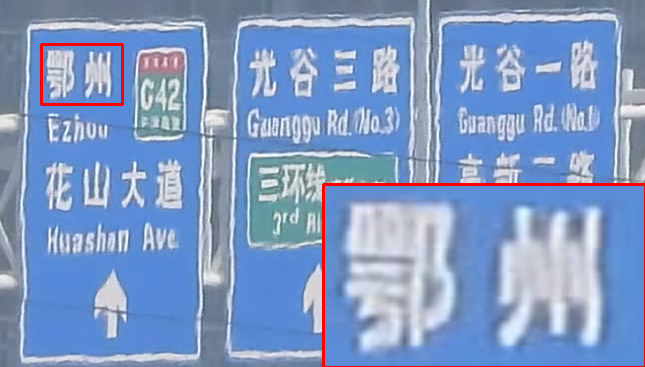}
        \hfill
        \includegraphics[height=3.7cm]{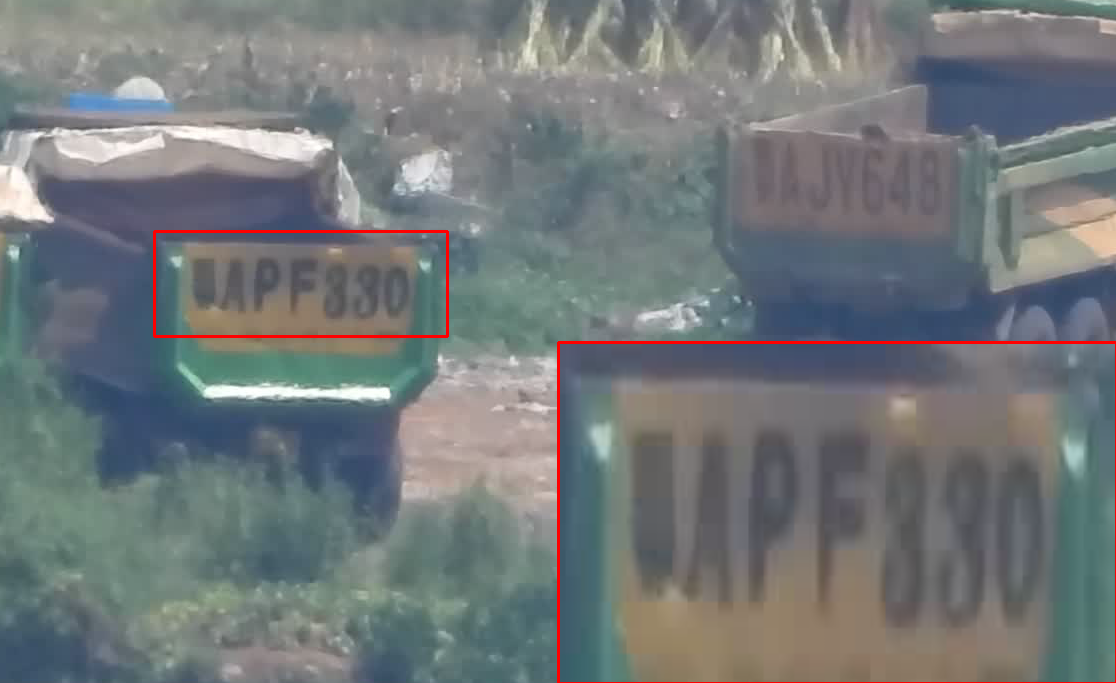}
        \caption{DATUM~\cite{zhang2024spatio}}
    \end{subfigure}
    % \vspace{2pt}
    
    % ===== Row 5: MambaTM =====
    \begin{subfigure}[b]{0.73\textwidth}
        \centering
        \includegraphics[height=3.7cm]{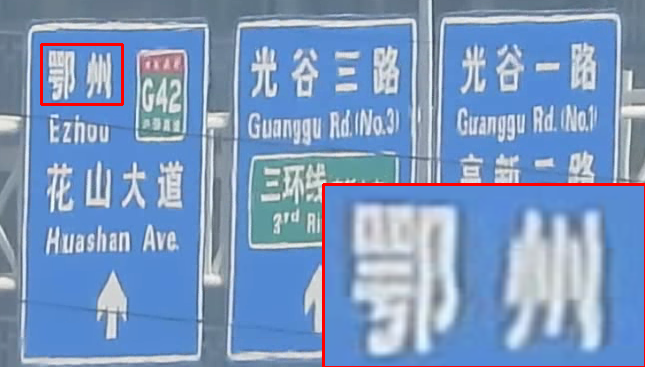}
        \hfill
        \includegraphics[height=3.7cm]{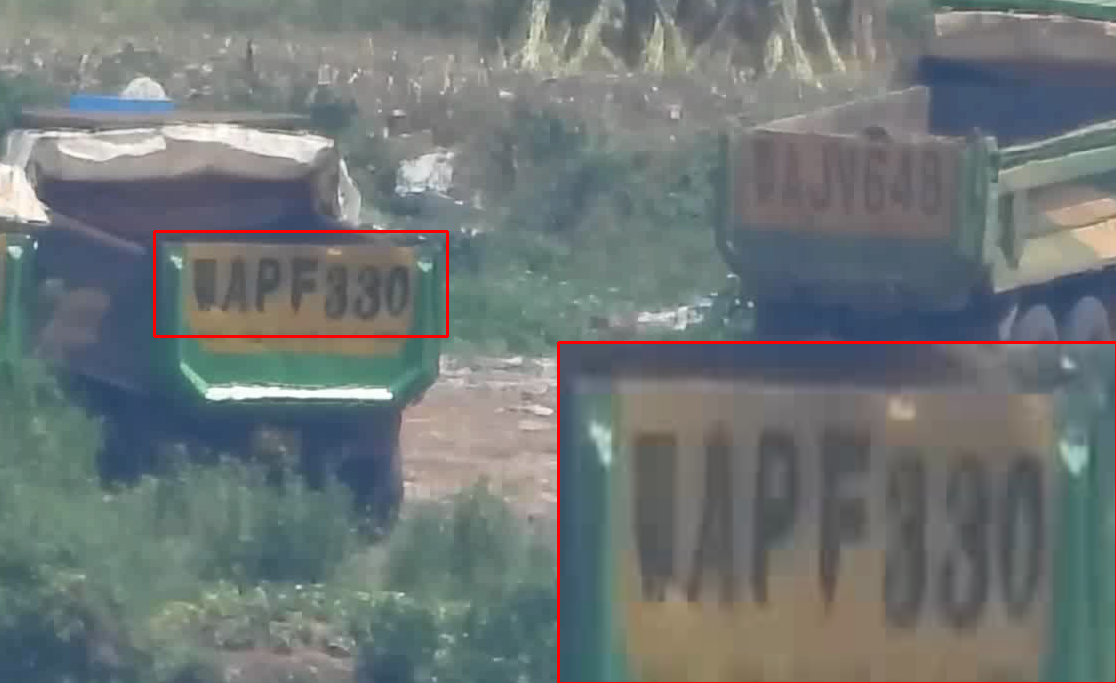}
        \caption{MambaTM~\cite{zhang2025learning}}
    \end{subfigure}
    % \vspace{2pt}
    
    \caption{Real-world turbulence mitigation results. Models trained on ET-Turb dataset demonstrate strong generalization to real atmospheric turbulence, successfully restoring fine details such as text and license plates while maintaining varying degrees of color fidelity.}
    \label{supp:fig_et_turb_real_comparison}
\end{figure*}

To evaluate the practical applicability of our ET-Turb dataset, we evaluate state-of-the-art turbulence mitigation algorithms trained on our synthetic data when applied to real-world turbulent imagery. Fig.~\ref{supp:fig_et_turb_real_comparison} presents qualitative results from four representative methods: TSR-WGAN~\cite{jin2021neutralizing}, TMT~\cite{zhang2024imaging}, DATUM~\cite{zhang2024spatio}, and MambaTM~\cite{zhang2025learning}.

The results demonstrate that models trained exclusively on ET-Turb successfully generalize to real atmospheric turbulence conditions. All evaluated algorithms achieve meaningful restoration of details, such as text on road signs and vehicle license plates, which are typically severely degraded by turbulence effects. Notably, while TSR-WGAN~\cite{jin2021neutralizing} produces relatively realistic detail recovery, it exhibits color shift artifacts compared to other methods. TMT~\cite{zhang2024imaging}, DATUM~\cite{zhang2024spatio}, and MambaTM~\cite{zhang2025learning} maintain better color fidelity while providing competitive detail restoration.

\begin{table}[t]
\centering
\caption{Quantitative evaluation on the ET-Turb test set. TMT~\cite{zhang2024imaging} achieves the best performance across all metrics, highlighting the effectiveness of its component-wise training strategy.}
\begin{tabular}{lccc}
\hline
Method   & PSNR(dB) $\uparrow$  & SSIM $\uparrow$ & LPIPS $\downarrow$  \\ \hline
TSR-WGAN~\cite{jin2021neutralizing} & 18.90 & 0.5927 & 0.5608 \\
TMT~\cite{zhang2024imaging}      & \textbf{24.97} & \textbf{0.7507} & \textbf{0.3807} \\
DATUM~\cite{zhang2024spatio}    & 22.85 & 0.6436 & 0.4700 \\
MambaTM~\cite{zhang2025learning}  & 23.77 & 0.6658 & 0.4490 \\ \hline
\end{tabular}
\label{tab:algorithm_benchmarking}
\end{table}

\begin{table*}[ht]
\centering
\caption{Robustness evaluation under exposure-dependent Gaussian noise added to the ET-Turb test set. Models trained on ET-Turb and ATSyn-Dynamic~\cite{zhang2024spatio} are compared.}
\label{tab:Robustness}
\scriptsize
\setlength{\tabcolsep}{2.2pt}
\renewcommand{\arraystretch}{0.95}
\resizebox{\linewidth}{!}{%
\begin{tabular}{lccc ccc}
\toprule
& \multicolumn{3}{c}{Trained on ET-Turb} 
& \multicolumn{3}{c}{Trained on ATSyn-Dynamic~\cite{zhang2024spatio}} \\
\cmidrule(lr){2-4}\cmidrule(lr){5-7}
Method & PSNR(dB)$\uparrow$ & SSIM$\uparrow$ & LPIPS$\downarrow$
       & PSNR(dB)$\uparrow$ & SSIM$\uparrow$ & LPIPS$\downarrow$ \\
\midrule
Noise-input   
& 20.57 (↓1.36\%) & 0.58 (↓9.38\%) & 0.57 (↑16.33\%) 
& 20.57 (↓1.36\%) & 0.58 (↓9.38\%) &0.57 (↑16.33\%) \\

TMT~\cite{zhang2024imaging}     
& 21.22 (↓0.38\%) & 0.66 (↓1.49\%) & 0.51 (↑2.21\%) 
& 21.19 (↓0.42\%) & 0.66 (↓2.94\%) & 0.51 (↑2.37\%) \\

DATUM~\cite{zhang2024spatio}   
& 22.43 (↓0.80\%) & 0.61 (↓4.69\%) & 0.51 (↑2.89\%) 
& 22.14 (↓1.07\%) & 0.62 (↓7.46\%) & 0.53 (↑8.16\%) \\

MambaTM~\cite{zhang2025learning} 
& 22.12 (↓1.07\%) & 0.64 (↓1.54\%) & 0.52 (↑4.49\%) 
& 22.10 (↓1.34\%) & 0.63 (↓4.55\%) & 0.52 (↑8.33\%) \\
\bottomrule
\end{tabular}}
\end{table*}

\begin{table*}[ht]
\centering
\caption{Sensitivity analyses with respect to $r_0$ and wind speed. The Fried parameter $r_0$ is perturbed by $\pm$10\%, while wind speed is increased from the training range (1--10\,m/s) to 10--20\,m/s.}
\label{tab:sensitivity_analyses}
\scriptsize
\setlength{\tabcolsep}{2.2pt}
\renewcommand{\arraystretch}{0.95}
\resizebox{\linewidth}{!}{%
\begin{tabular}{lccc ccc}
\toprule
& \multicolumn{3}{c}{$r_0$ ±10\%} 
& \multicolumn{3}{c}{wind speed within 10-20m/s} \\
\cmidrule(lr){2-4}\cmidrule(lr){5-7}
Method & PSNR(dB)$\uparrow$ & SSIM$\uparrow$ & LPIPS$\downarrow$
       & PSNR(dB)$\uparrow$ & SSIM$\uparrow$ & LPIPS$\downarrow$ \\
\midrule
Input   
& 20.71 (↓0.07\%) & 0.78 (↓0.14\%) & 0.34 (↑0.25\%) 
& 20.43 (↓0.51\%) & 0.76 (↓2.07\%) &0.36 (↑1.88\%) \\

TMT~\cite{zhang2024imaging}     
& 21.20 (↓0.06\%) & 0.83 (↓0.14\%) & 0.33 (↑0.12\%) 
& 20.83 (↓0.82\%) & 0.80 (↓1.19\%) & 0.36 (↑2.22\%) \\

DATUM~\cite{zhang2024spatio}   
& 22.27 (↓0.11\%) & 0.79 (↓0.17\%) & 0.33 (↑0.18\%) 
& 21.75 (↓0.87\%) & 0.76 (↓1.48\%) & 0.36 (↑2.37\%) \\

MambaTM~\cite{zhang2025learning} 
& 21.57 (↓0.06\%) & 0.78 (↓0.13\%) & 0.36 (↑0.00\%) 
& 21.19 (↓0.60\%) & 0.75 (↓1.23\%) & 0.38 (↑1.99\%) \\
\bottomrule
\end{tabular}}
\end{table*}

\subsection{Qualitative and Quantitative Evaluation on ET-Turb Test Set}
Tab.~\ref{tab:algorithm_benchmarking} presents a comprehensive quantitative comparison of four algorithms evaluated on the ET-Turb test set. Performance is assessed using three complementary metrics: PSNR for pixel-level accuracy, SSIM for structural similarity, and LPIPS~\cite{zhang2018unreasonable} for perceptual quality assessment.

Among the algorithms tested, TMT~\cite{zhang2024imaging} achieves the best overall performance across all metrics (PSNR: 24.97 dB, SSIM: 0.7507, LPIPS: 0.3807), due to its approach of separately training the tilt and blur components. MambaTM~\cite{zhang2025learning} follows closely with balanced results, then followed by DATUM~\cite{zhang2024spatio}. TSR-WGAN~\cite{jin2021neutralizing} shows a notable performance gap compared with other methods.

\begin{figure*}[p]
    \centering

    % ===== Row 1: Input =====
    \begin{subfigure}[b]{0.65\textwidth}
        \centering
        \includegraphics[width=0.48\textwidth]{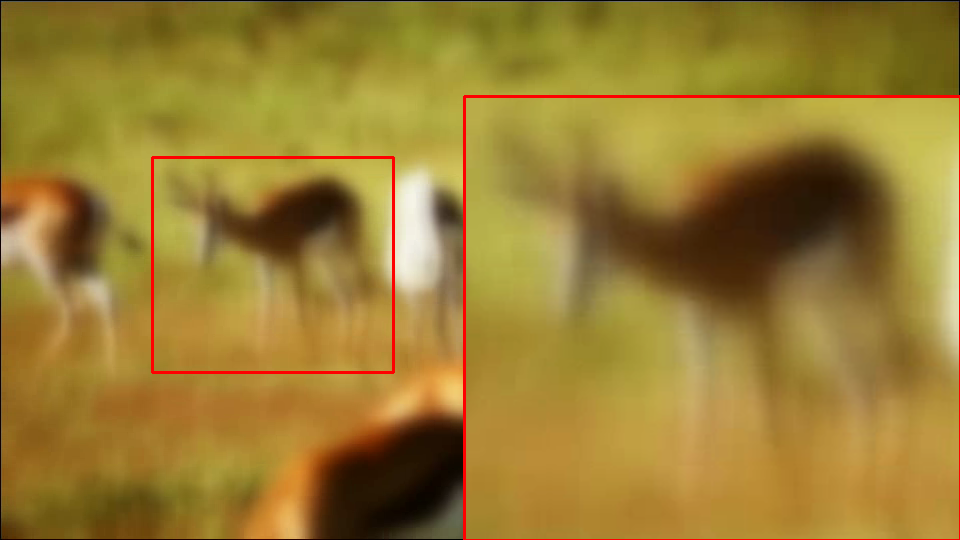}
        \hfill
        \includegraphics[width=0.48\textwidth]{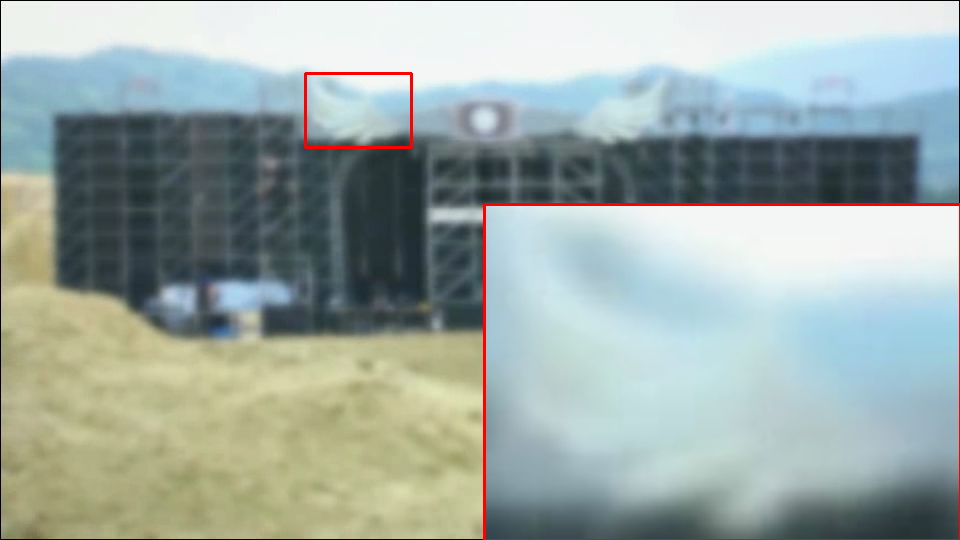}
        \caption{Input}
    \end{subfigure}

    % \vspace{2pt}

    % ===== Row 2: TSR-WGAN =====
    \begin{subfigure}[b]{0.65\textwidth}
        \centering
        \includegraphics[width=0.48\textwidth]{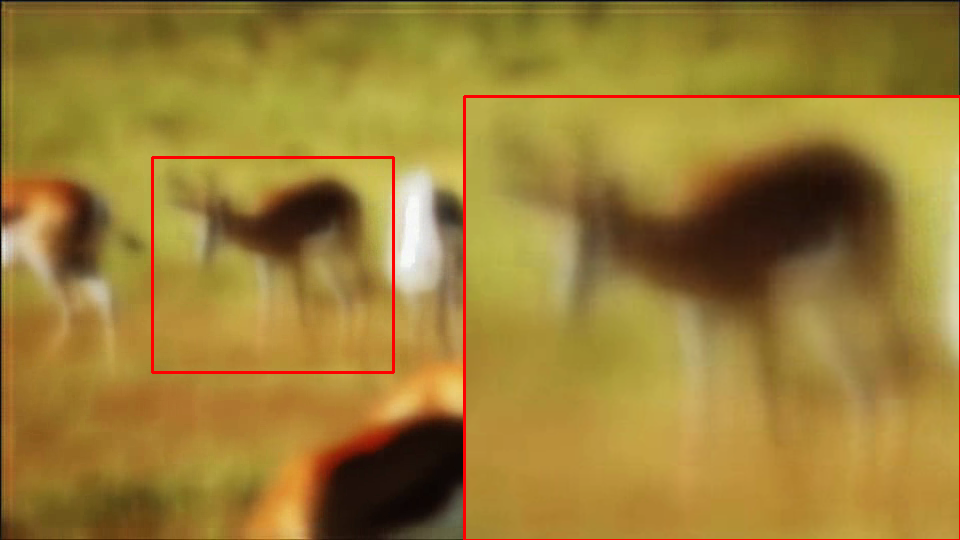}
        \hfill
        \includegraphics[width=0.48\textwidth]{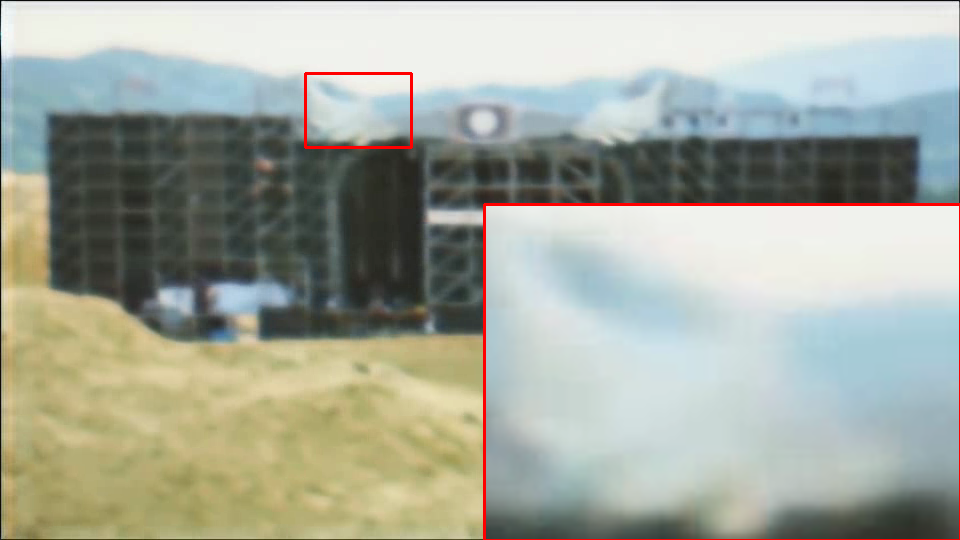}
        \caption{TSR-WGAN~\cite{jin2021neutralizing}}
    \end{subfigure}

    % \vspace{2pt}

    % ===== Row 3: TMT =====
    \begin{subfigure}[b]{0.65\textwidth}
        \centering
        \includegraphics[width=0.48\textwidth]{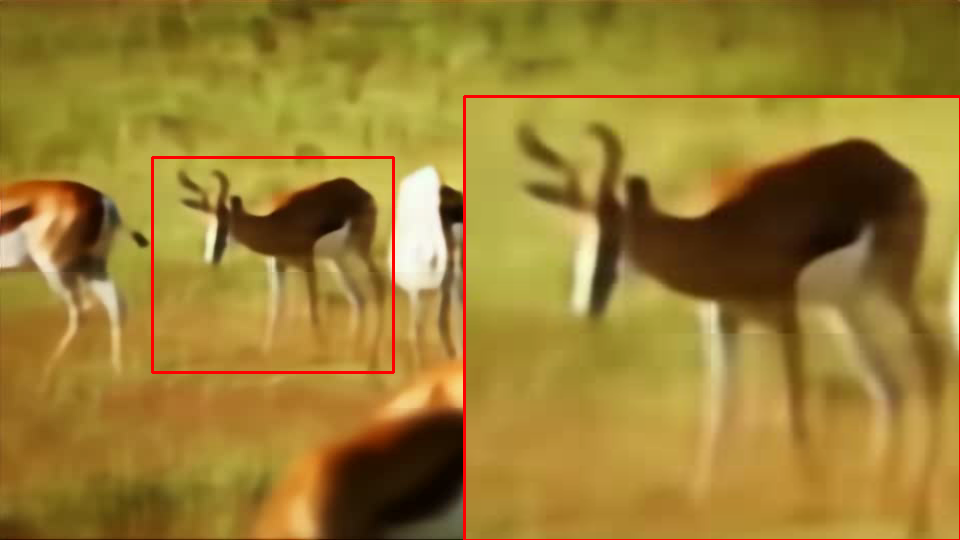}
        \hfill
        \includegraphics[width=0.48\textwidth]{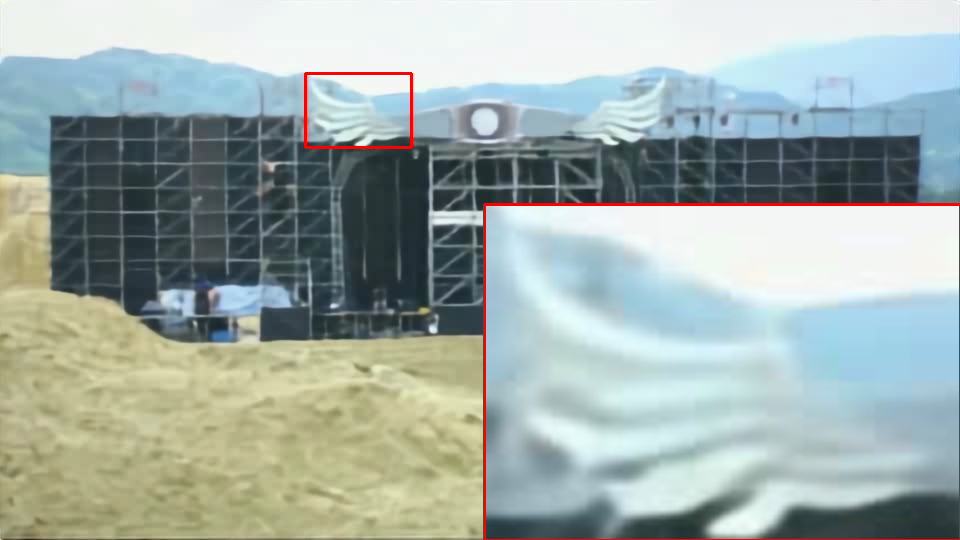}
        \caption{TMT~\cite{zhang2024imaging}}
    \end{subfigure}

    % \vspace{2pt}

    % ===== Row 4: DATUM =====
    \begin{subfigure}[b]{0.65\textwidth}
        \centering
        \includegraphics[width=0.48\textwidth]{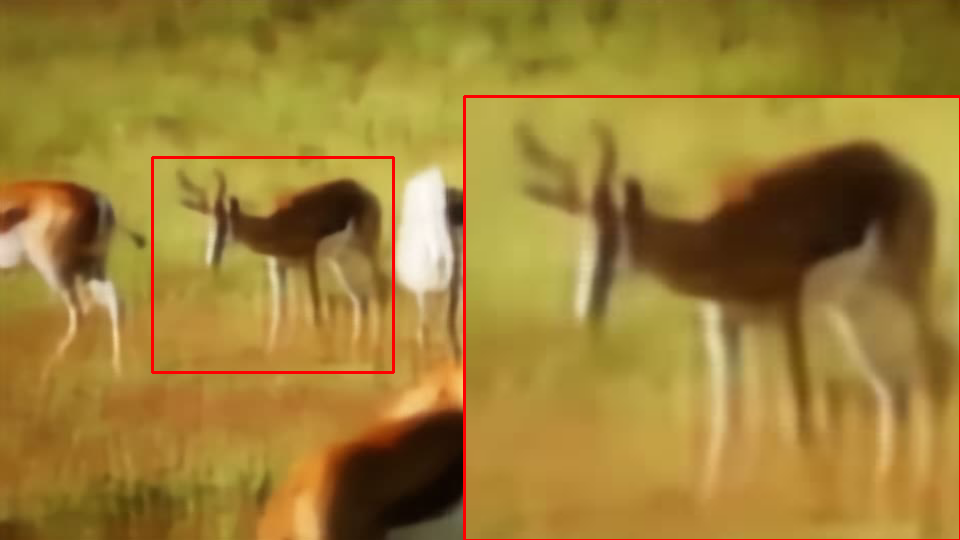}
        \hfill
        \includegraphics[width=0.48\textwidth]{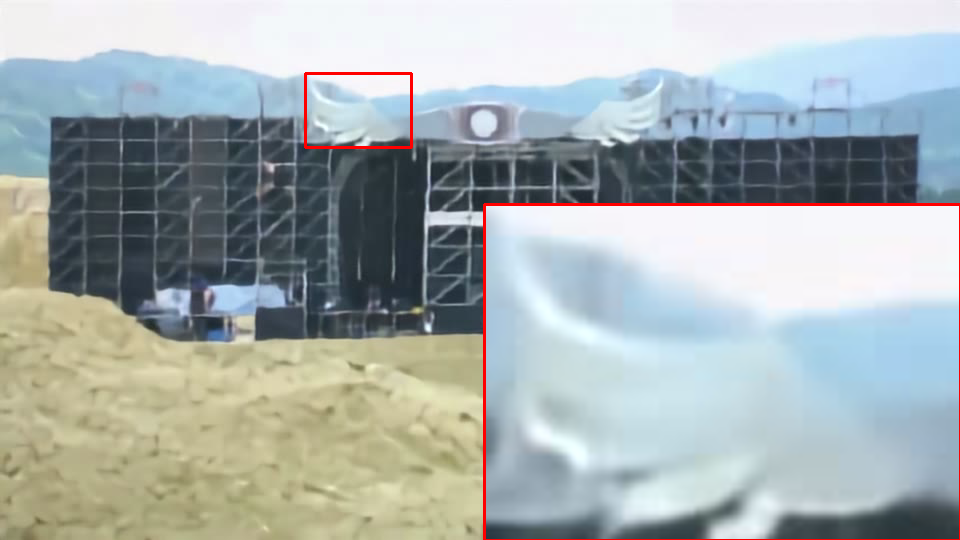}
        \caption{DATUM~\cite{zhang2024spatio}}
    \end{subfigure}

    % \vspace{2pt}

    % ===== Row 5: MambaTM =====
    \begin{subfigure}[b]{0.65\textwidth}
        \centering
        \includegraphics[width=0.48\textwidth]{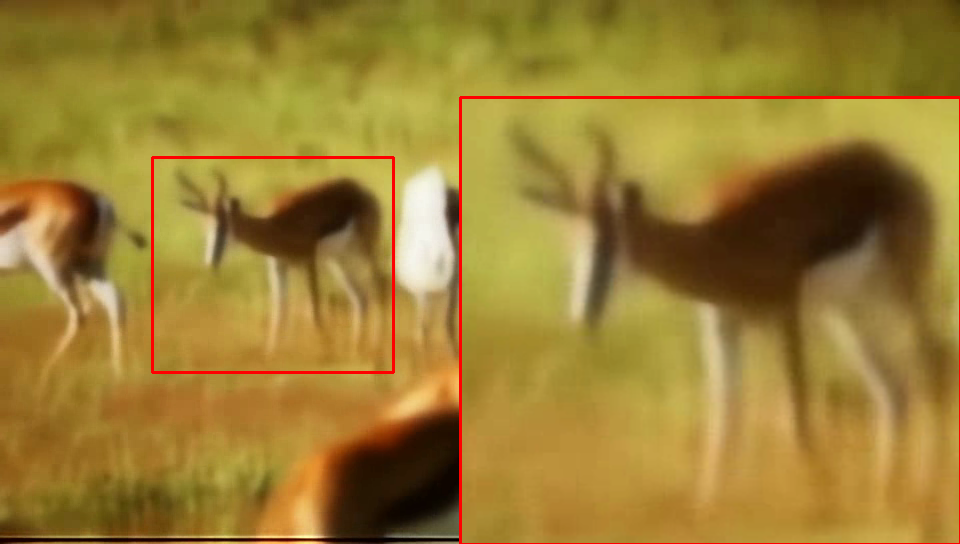}
        \hfill
        \includegraphics[width=0.48\textwidth]{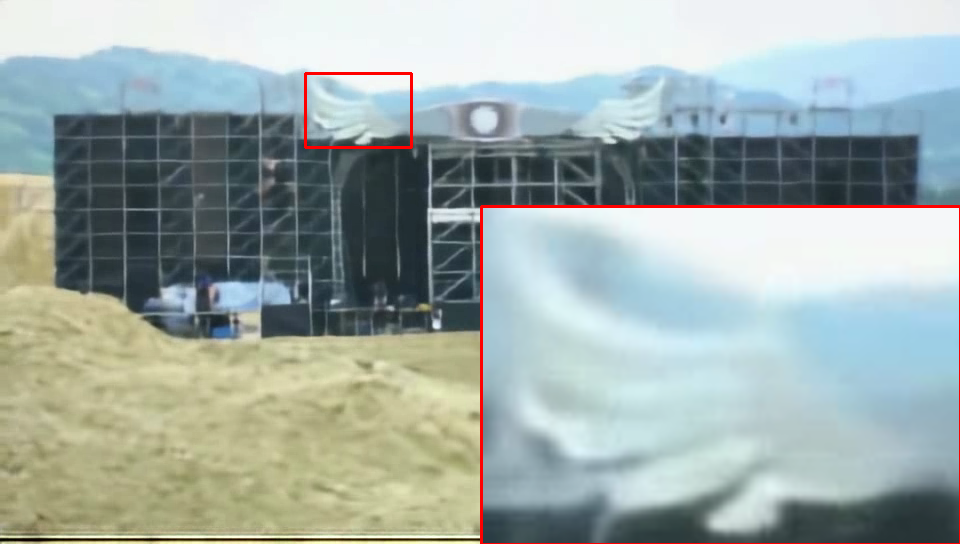}
        \caption{MambaTM~\cite{zhang2025learning}}
    \end{subfigure}

    % ===== Row 6: GT =====
    \begin{subfigure}[b]{0.65\textwidth}
        \centering
        \includegraphics[width=0.48\textwidth]{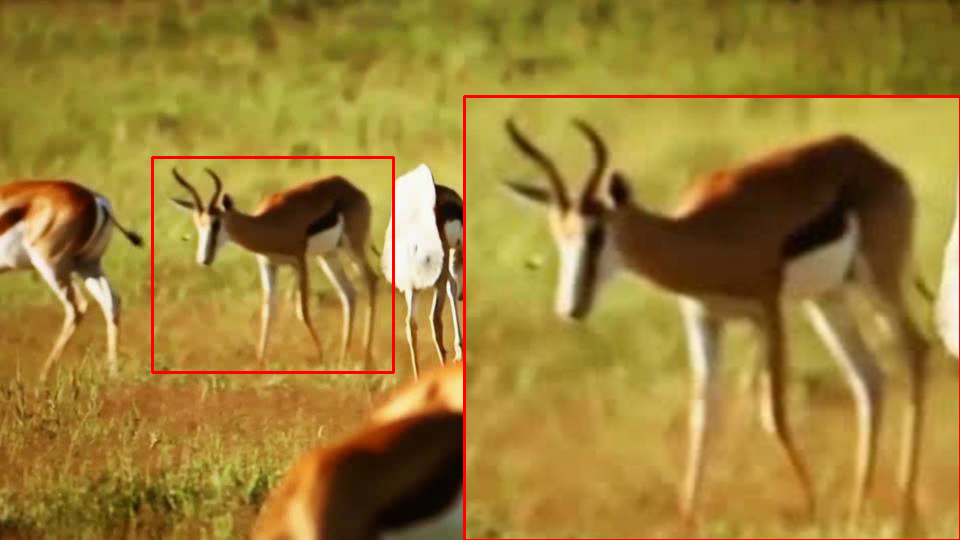}
        \hfill
        \includegraphics[width=0.48\textwidth]{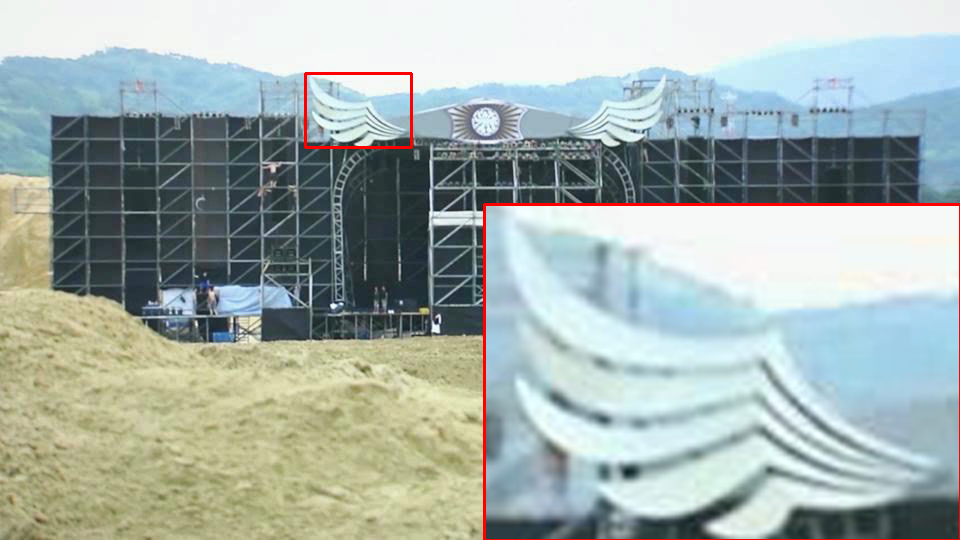}
        \caption{GT}
    \end{subfigure}

    \caption{Qualitative comparison of state-of-the-art turbulence mitigation methods trained on ET-Turb dataset. Each row corresponds to a specific method, showing results from two representative test scenes (left and right).}
    \label{supp:fig_algorithm_benchmarking}
\end{figure*}

In addition to the quantitative comparison, qualitative assessments, shown in Fig.~\ref{supp:fig_algorithm_benchmarking}, corroborate the numerical results, with TMT producing the most visually accurate images. However, TSR-WGAN exhibits visible artifacts.

These results validate the effectiveness of the ET-Turb dataset for training turbulence mitigation algorithms and suggest that existing methods can be improved, particularly with more realistic turbulence datasets.

\begin{figure}[t]
  \centering
\includegraphics[width=\columnwidth]{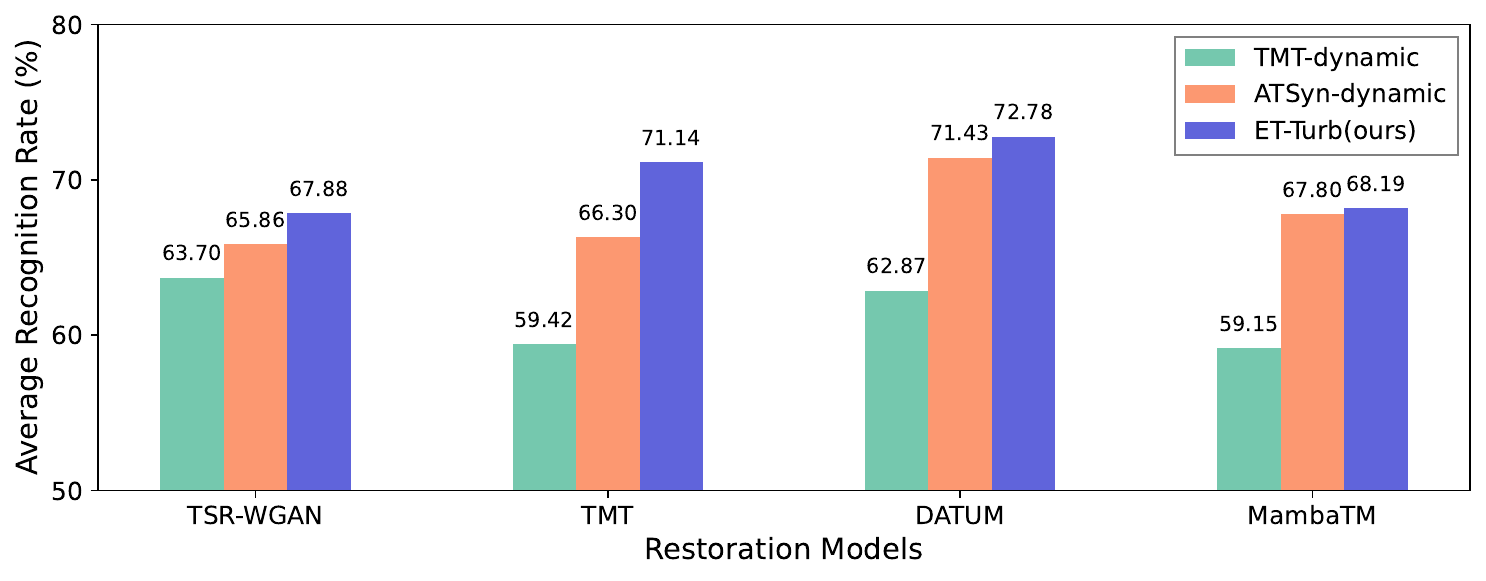} 
  \caption{Text recognition accuracy on the turbulence-text dataset~\cite{zhang2024imaging}, evaluated using ASTER~\cite{shi2018aster}.}
   % \caption{Text recognition test on Turbulence-Text dataset.}
  \label{fig:text} 
\end{figure}

\subsection{Text Recognition Evaluation}

To evaluate the downstream utility of the synthesized turbulence data, we conduct a text recognition experiment on the Turbulence-Text dataset~\cite{zhang2024imaging} using ASTER~\cite{shi2018aster} as the recognition backbone and measure accuracy on restored frames.

As shown in Fig.~\ref{fig:text}, models trained on ET-Turb achieve superior recognition performance compared to those trained on TMT-dynamic~\cite{zhang2024imaging} and ATSyn-dynamic~\cite{zhang2024spatio}. In particular, ET-Turb improves recognition accuracy by 7.21\% over TMT-dynamic~\cite{zhang2024imaging} and 1.15\% over ATSyn-dynamic~\cite{zhang2024spatio}, indicating better downstream generalization.

\subsection{Exposure-Time Decoupling and Brightness Statistics}

In practical imaging systems, exposure-time $\tau$, brightness, and sensor noise are physically coupled. However, the primary objective of this work is turbulence synthesis under continuous exposure modeling, rather than full radiometric simulation. In real-world surveillance scenarios, cameras typically operate with auto-exposure mechanisms, where $\tau$ does not uniquely determine image brightness due to ISO and aperture adjustments that vary across camera designs. Explicitly modeling these factors would introduce additional parameters beyond the scope of this work.

\begin{figure}[t]
  \centering
  \includegraphics[width=\columnwidth]{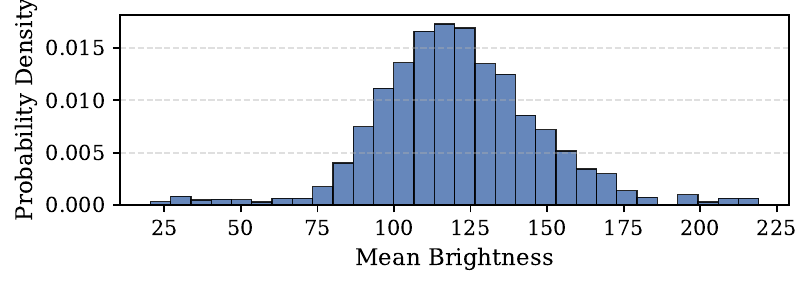} 
  \caption{Brightness distribution statistics of ET-Turb.}
  \label{fig:total} 
\end{figure}

Although brightness is not explicitly coupled with $\tau$ in our formulation, ET-Turb spans a broad brightness range. As shown in Fig.~\ref{fig:total}, the brightness distribution follows a smooth normal-like distribution, ensuring sufficient radiometric diversity for model training.

\subsection{Robustness to Exposure-Dependent Noise}

We introduce exposure-dependent Gaussian noise defined as $\sigma(\tau)=K/\sqrt{\tau}$ with $K=0.001$ and $\tau \in [0.5, 40]$\,ms to evaluate robustness under radiometric perturbations. Noise is added to a randomly selected subset (10\%) of the ET-Turb test set (100 videos), then models trained on ET-Turb and ATSyn-Dynamic~\cite{zhang2024spatio} are evaluated.

As summarized in Tab.~\ref{tab:Robustness}, all models exhibit performance degradation under noise. However, models trained on ET-Turb demonstrate consistently smaller performance drops compared to those trained on ATSyn-Dynamic~\cite{zhang2024spatio}, indicating improved robustness to radiometric perturbations compared to ATSyn-Dynamic~\cite{zhang2024spatio}.

\subsection{Sensitivity Analyses}
To further evaluate the robustness of models trained on ET-Turb, we conduct additional sensitivity analyses with respect to key atmospheric parameters, including the Fried parameter $r_0$ and wind speed. Specifically, $r_0$ is perturbed by $\pm$10\%, and wind speed is varied from the training range (1-10 m/s) to an unseen higher range (10-20 m/s). The results in Tab.~\ref{tab:sensitivity_analyses} demonstrate stable performance under both parameter variations, with only marginal degradation across all evaluation metrics. This indicates the robustness of models trained on ET-Turb to atmospheric parameter shifts.

\begin{table}[t]
\centering
\caption{Processing time for a $256^2$ frame.}
\label{tab:computational_cost}
\begin{tabular}{ccclll}
\hline
Methods & Hardie & Chimitt & Lau & P2S & Ours \\
\hline
Sec./frame & 119.63 & 5.88 & 3.13 & 0.35 & 3.24 \\
\hline
\end{tabular}
\end{table}

\subsection{Computational Cost}

We compare the run time of the proposed method with several representative approaches~\cite{hardie2017simulation, chimitt2020simulating, lau2021atfacegan, mao2021accelerating}. The processing time for a $256^2$ frame is summarized in Tab.~\ref{tab:computational_cost}, where the runtimes of competing methods are taken from~\cite{mao2021accelerating}. Our method requires 3.24\,s/frame, which is comparable to Lau~\cite{lau2021atfacegan} (3.13\,s) but slower than P2S~\cite{mao2021accelerating} (0.35\,s). When accelerated with GPU, our method consumes 448\,MB.

\begin{figure*}[p]
  \centering

  \captionsetup{type=table} 
    \begin{minipage}{0.75\textwidth} 
      \centering
      \caption{Physical and optical parameters for the six turbulence synthesis configurations. These parameters span realistic ranges encountered in practical atmospheric imaging scenarios.}
      \setlength{\tabcolsep}{4pt}  
      \renewcommand{\arraystretch}{1.05} 
      \footnotesize                   
    
      \begin{tabular*}{\linewidth}{@{\extracolsep{\fill}} c c c c c c c}
        \hline
        Distance (m) & Focal length (m) & F-number &
        \begin{tabular}{@{}c@{}}$C_n^2$\\ ($10^{-14}\,\mathrm{m}^{-2/3}$)\end{tabular} &
        Height (m) & Wind speed (m/s) &
        \begin{tabular}{@{}c@{}}Exposure\\ time (ms)\end{tabular} \\ \hline
        50  & 0.1 & 4   & 260 & 4   & 2 & 4  \\ \hline
        328 & 0.4 & 5.6 & 16  & 50  & 2 & 10 \\ \hline
        339 & 0.3 & 11  & 29  & 10  & 5 & 20 \\ \hline
        518 & 0.5 & 8   & 10  & 50  & 5 & 8  \\ \hline
        517 & 0.5 & 11  & 17  & 10  & 5 & 10 \\ \hline
        724 & 0.8 & 8   & 11  & 100 & 4 & 20 \\ \hline
      \end{tabular*}
    
      \label{supp:parameter_range}
    \end{minipage}

  \vspace{8pt}
  \captionsetup{type=figure} 

  % Row 1: GT + TILT + BLUR + TURB
  \begin{subfigure}[b]{0.245\textwidth}
    \includegraphics[width=\linewidth,height=2.4cm,keepaspectratio]{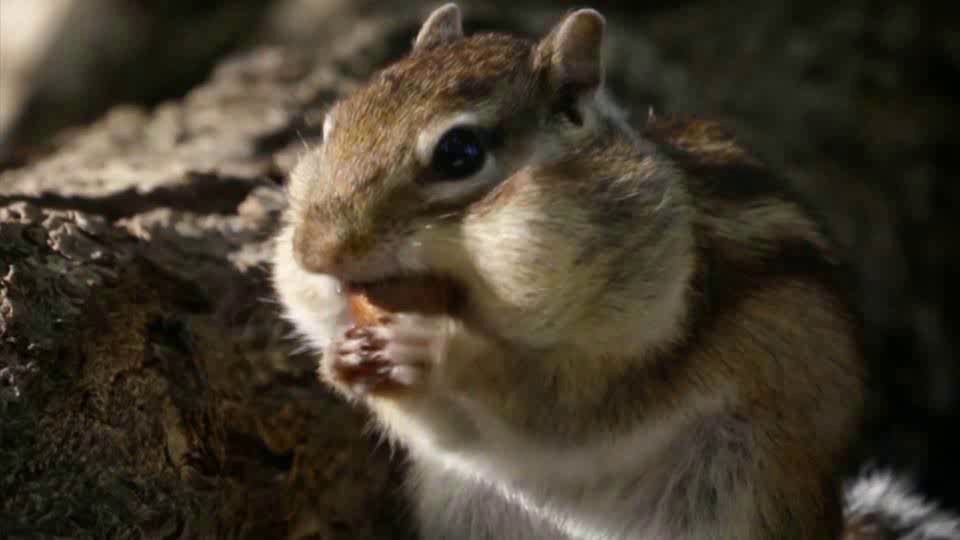}
  \end{subfigure}\hspace{3pt}%
  \begin{subfigure}[b]{0.245\textwidth}
    \includegraphics[width=\linewidth,height=2.4cm,keepaspectratio]{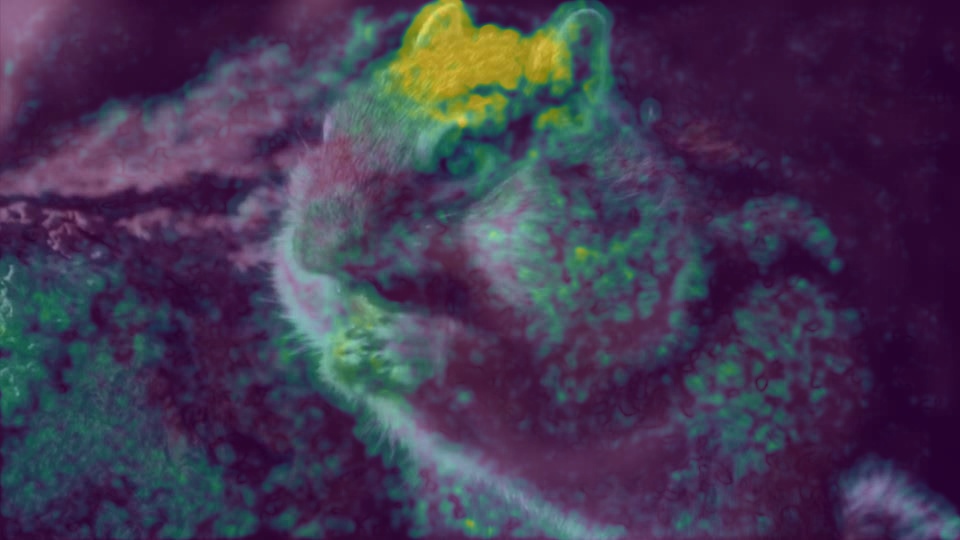}
  \end{subfigure}\hspace{3pt}%
  \begin{subfigure}[b]{0.245\textwidth}
    \includegraphics[width=\linewidth,height=2.4cm,keepaspectratio]{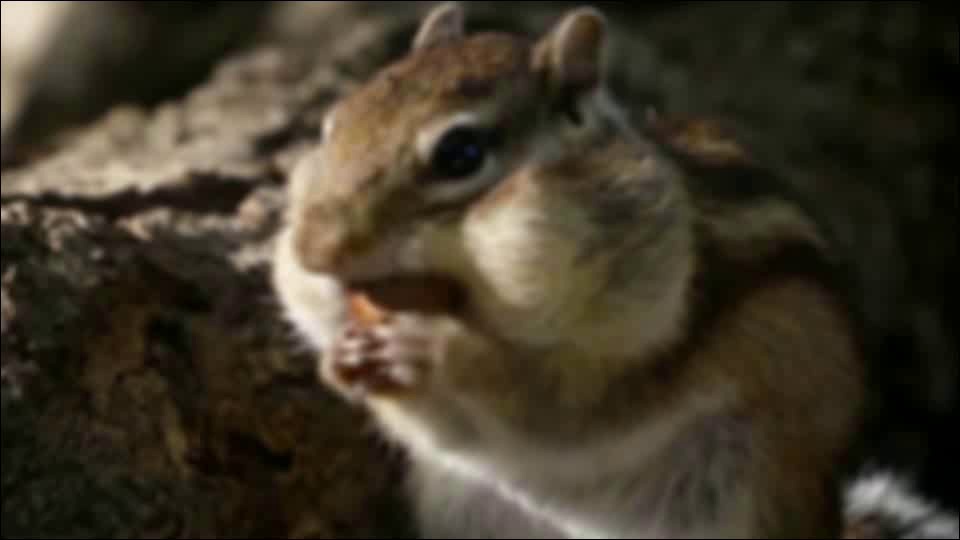}
  \end{subfigure}\hspace{3pt}%
  \begin{subfigure}[b]{0.245\textwidth}
    \includegraphics[width=\linewidth,height=2.4cm,keepaspectratio]{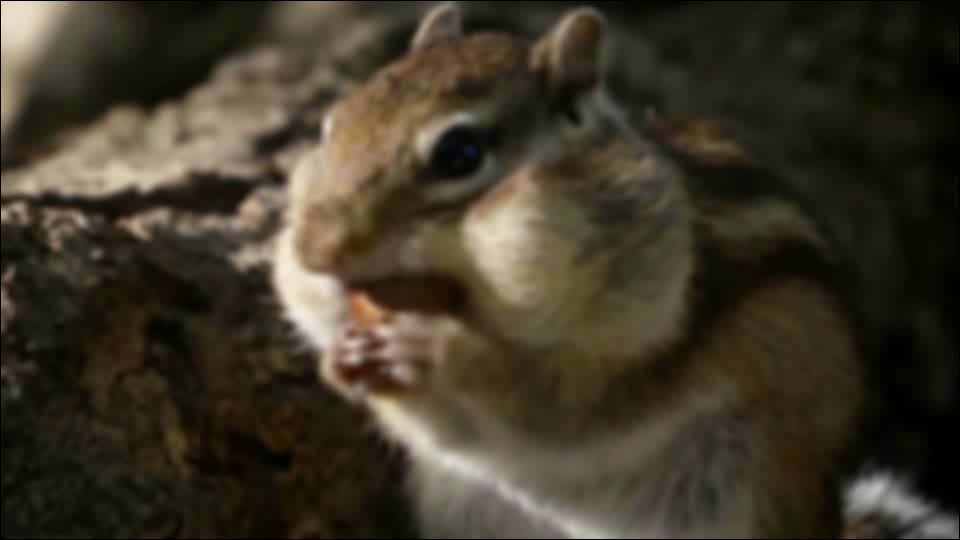}
  \end{subfigure}
  \par\vspace{4pt}

  % Row 2
  \begin{subfigure}[b]{0.245\textwidth}\rule{0pt}{2.4cm}\end{subfigure}\hspace{3pt}%
  \begin{subfigure}[b]{0.245\textwidth}\includegraphics[width=\linewidth,height=2.4cm,keepaspectratio]{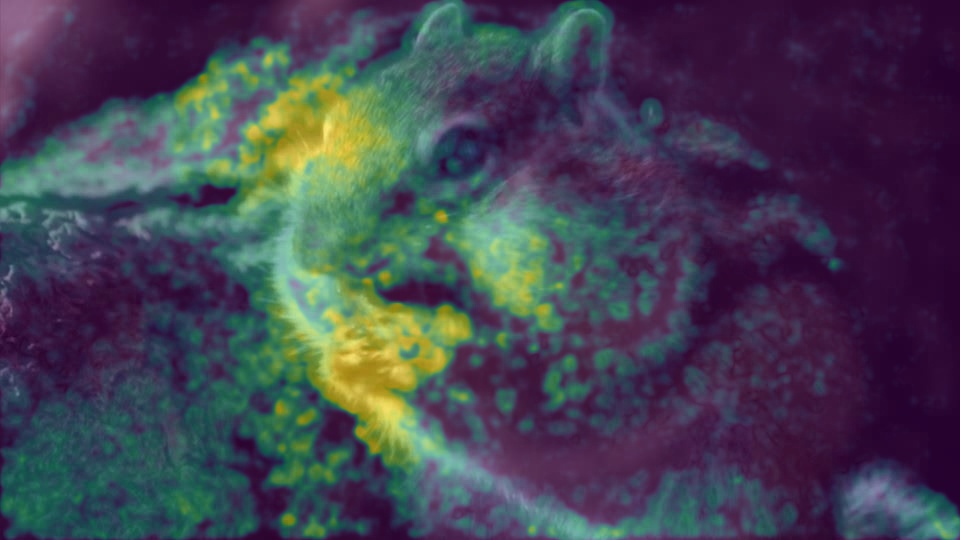}\end{subfigure}\hspace{3pt}%
  \begin{subfigure}[b]{0.245\textwidth}\includegraphics[width=\linewidth,height=2.4cm,keepaspectratio]{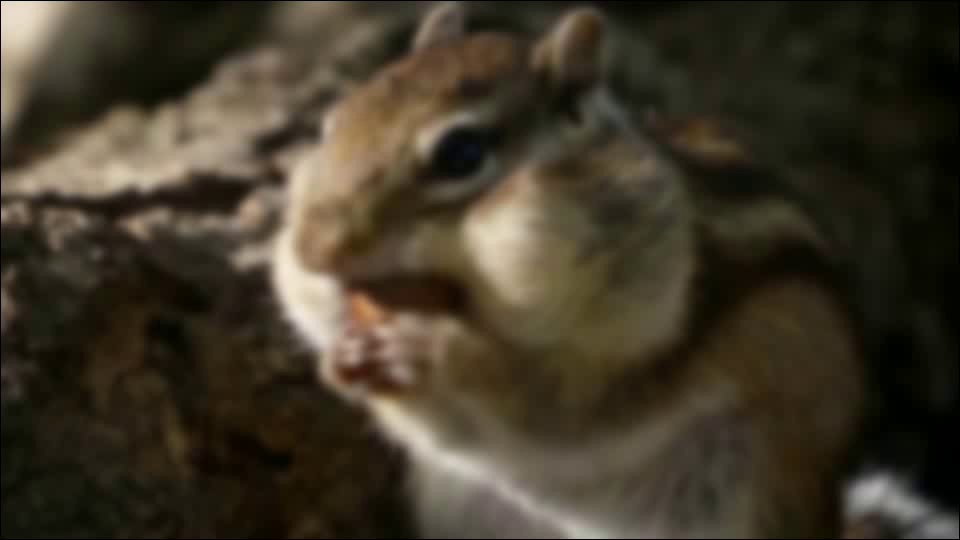}\end{subfigure}\hspace{3pt}%
  \begin{subfigure}[b]{0.245\textwidth}\includegraphics[width=\linewidth,height=2.4cm,keepaspectratio]{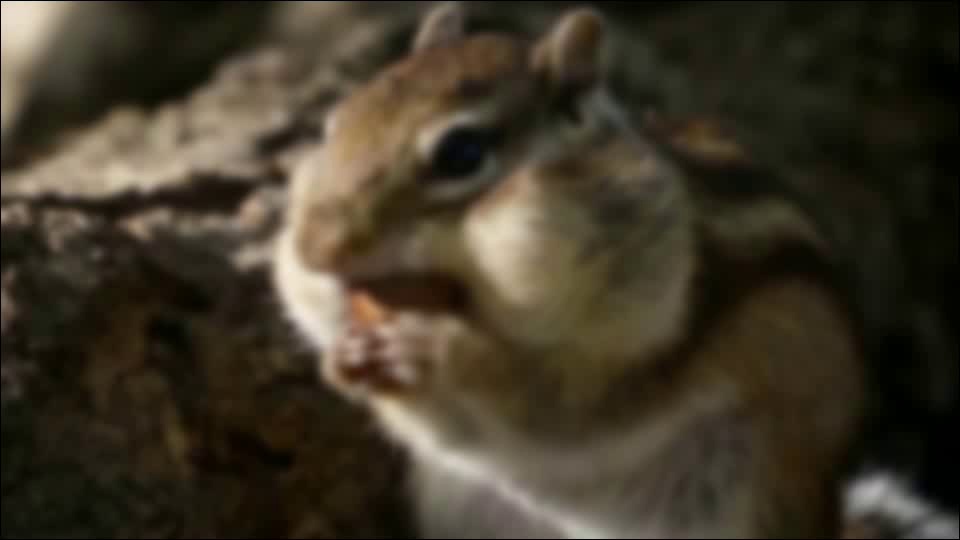}\end{subfigure}
  \par\vspace{4pt}

  % Row 3
  \begin{subfigure}[b]{0.245\textwidth}\rule{0pt}{2.4cm}\end{subfigure}\hspace{3pt}%
  \begin{subfigure}[b]{0.245\textwidth}\includegraphics[width=\linewidth,height=2.4cm,keepaspectratio]{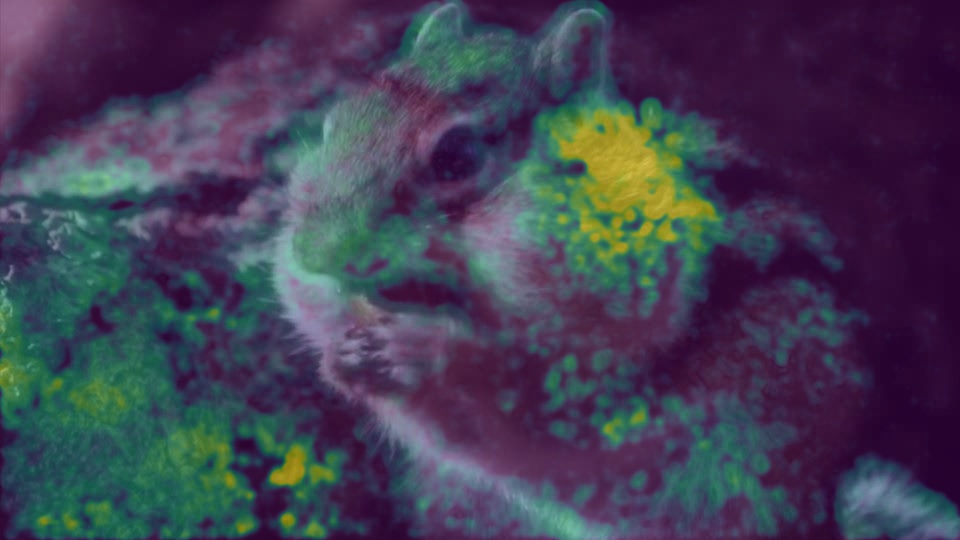}\end{subfigure}\hspace{3pt}%
  \begin{subfigure}[b]{0.245\textwidth}\includegraphics[width=\linewidth,height=2.4cm,keepaspectratio]{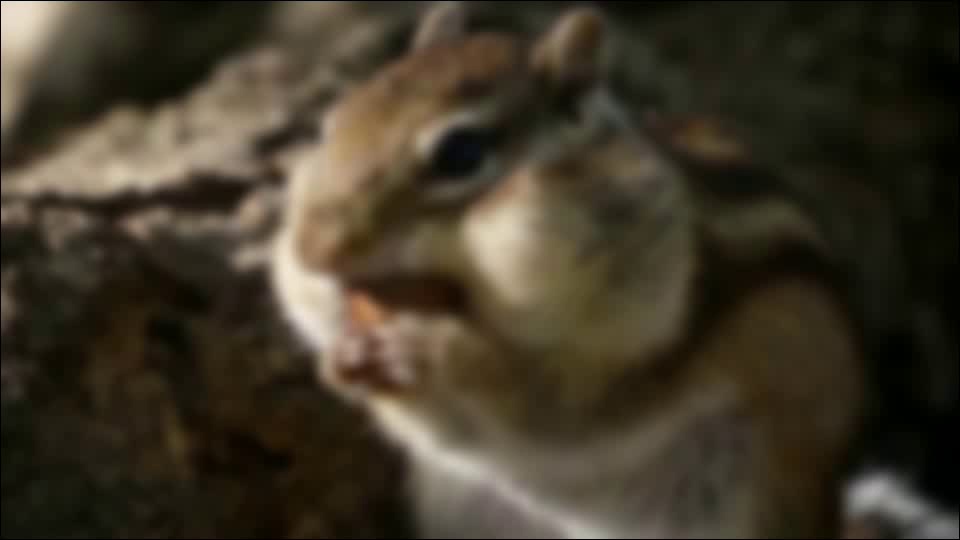}\end{subfigure}\hspace{3pt}%
  \begin{subfigure}[b]{0.245\textwidth}\includegraphics[width=\linewidth,height=2.4cm,keepaspectratio]{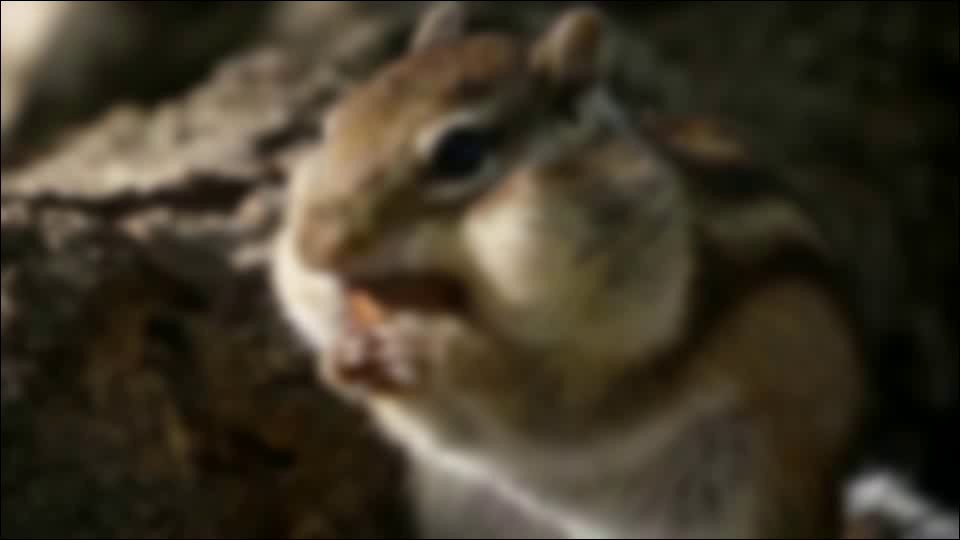}\end{subfigure}
  \par\vspace{4pt}

  % Row 4
  \begin{subfigure}[b]{0.245\textwidth}\rule{0pt}{2.4cm}\end{subfigure}\hspace{3pt}%
  \begin{subfigure}[b]{0.245\textwidth}\includegraphics[width=\linewidth,height=2.4cm,keepaspectratio]{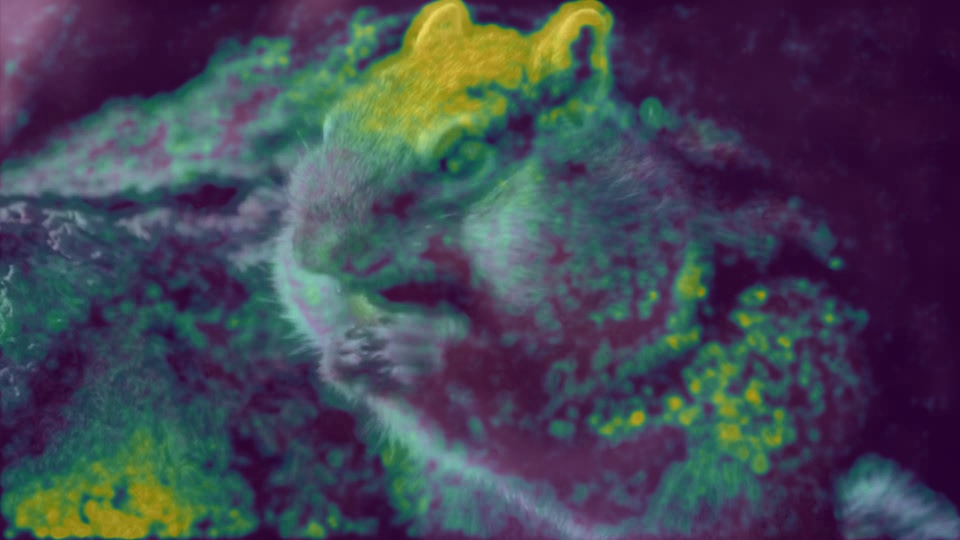}\end{subfigure}\hspace{3pt}%
  \begin{subfigure}[b]{0.245\textwidth}\includegraphics[width=\linewidth,height=2.4cm,keepaspectratio]{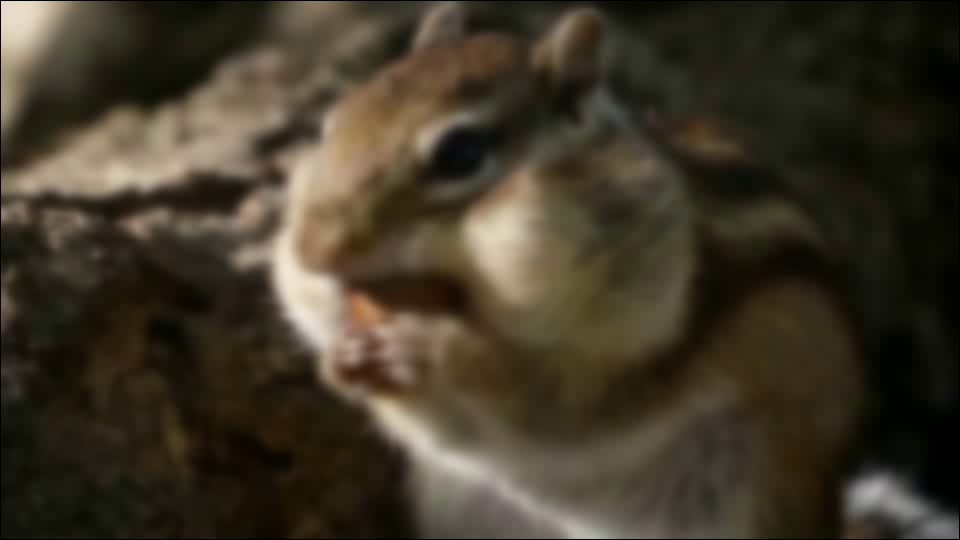}\end{subfigure}\hspace{3pt}%
  \begin{subfigure}[b]{0.245\textwidth}\includegraphics[width=\linewidth,height=2.4cm,keepaspectratio]{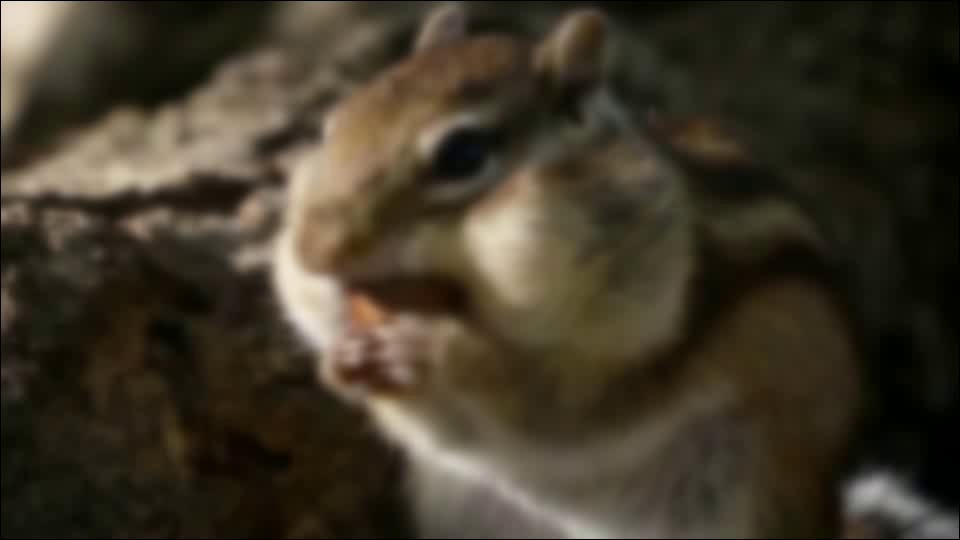}\end{subfigure}
  \par\vspace{4pt}

  % Row 5
  \begin{subfigure}[b]{0.245\textwidth}\rule{0pt}{2.4cm}\end{subfigure}\hspace{3pt}%
  \begin{subfigure}[b]{0.245\textwidth}\includegraphics[width=\linewidth,height=2.4cm,keepaspectratio]{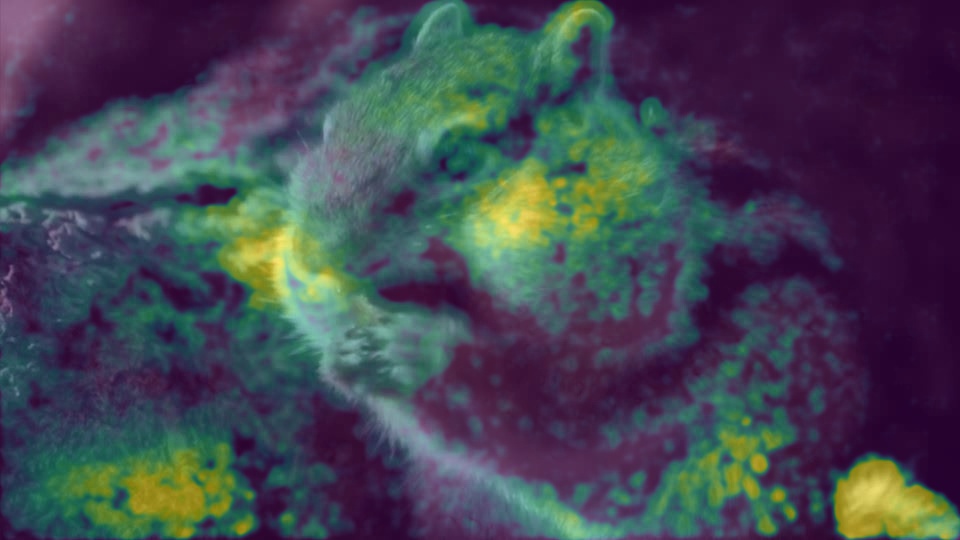}\end{subfigure}\hspace{3pt}%
  \begin{subfigure}[b]{0.245\textwidth}\includegraphics[width=\linewidth,height=2.4cm,keepaspectratio]{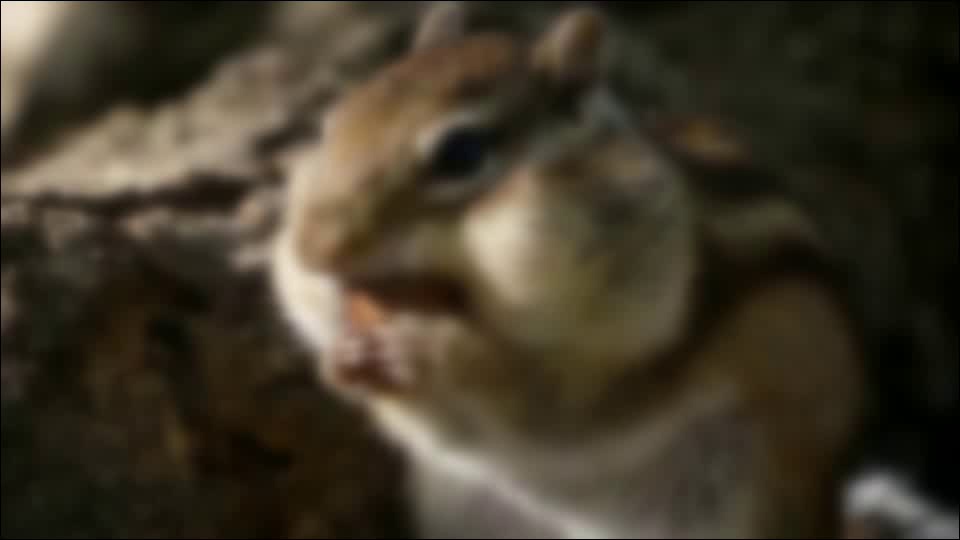}\end{subfigure}\hspace{3pt}%
  \begin{subfigure}[b]{0.245\textwidth}\includegraphics[width=\linewidth,height=2.4cm,keepaspectratio]{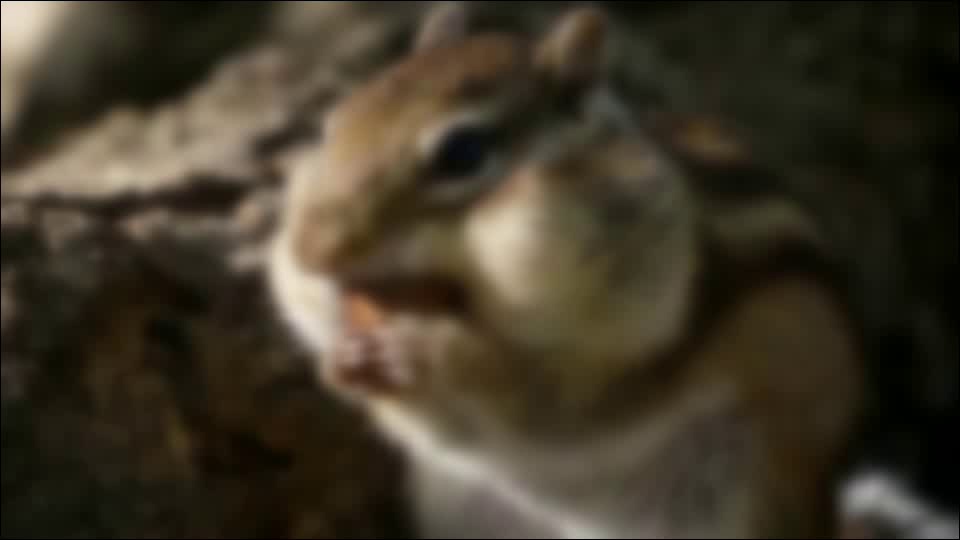}\end{subfigure}
  \par\vspace{4pt}

  % Row 6
  \begin{subfigure}[b]{0.245\textwidth}\rule{0pt}{2.4cm}\end{subfigure}\hspace{3pt}%
  \begin{subfigure}[b]{0.245\textwidth}\includegraphics[width=\linewidth,height=2.4cm,keepaspectratio]{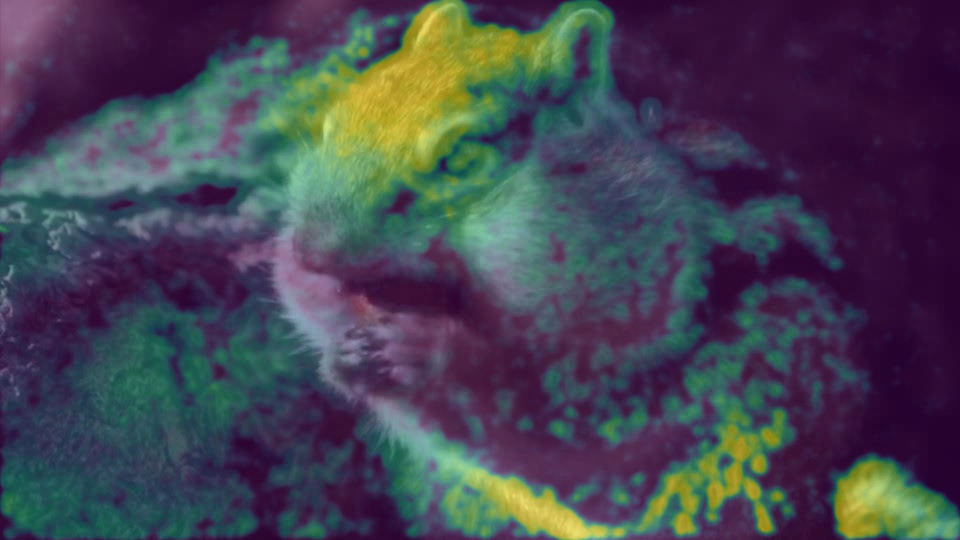}\end{subfigure}\hspace{3pt}%
  \begin{subfigure}[b]{0.245\textwidth}\includegraphics[width=\linewidth,height=2.4cm,keepaspectratio]{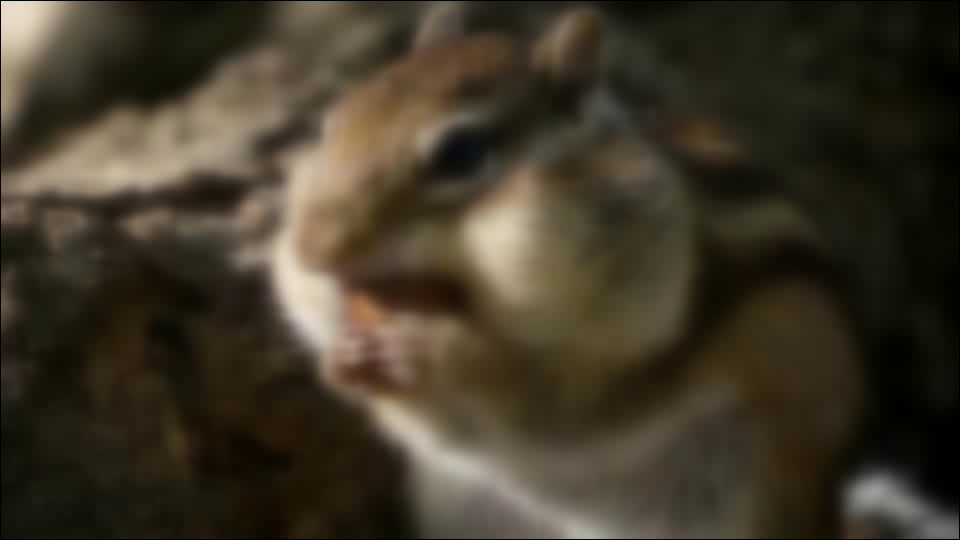}\end{subfigure}\hspace{3pt}%
  \begin{subfigure}[b]{0.245\textwidth}\includegraphics[width=\linewidth,height=2.4cm,keepaspectratio]{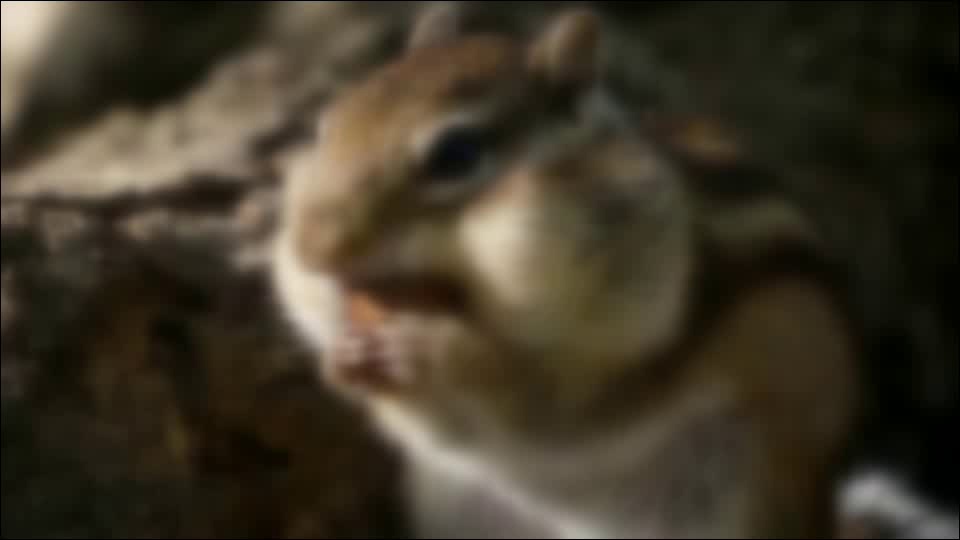}\end{subfigure}

  % Column labels as numbered subcaptions
  % \vspace{1pt}
  \captionsetup[subfigure]{labelformat=parens, labelsep=space, font=small, skip=2pt, justification=centering}
  \begin{subfigure}[t]{0.245\textwidth}\centering \caption{GT}\label{fig:col-gt}\end{subfigure}\hspace{3pt}%
  \begin{subfigure}[t]{0.245\textwidth}\centering \caption{Tilt}\label{fig:col-tilt}\end{subfigure}\hspace{3pt}%
  \begin{subfigure}[t]{0.245\textwidth}\centering \caption{Blur}\label{fig:col-blur}\end{subfigure}\hspace{3pt}%
  \begin{subfigure}[t]{0.245\textwidth}\centering \caption{Turb}\label{fig:col-turb}\end{subfigure}

  \caption{Visualization of the turbulence synthesis pipeline across six parameter configurations. Each row corresponds to a distinct set of physical parameters from Tab.~\ref{supp:parameter_range}. The columns show: (a) original undistorted image, (b) tilt with color-coded displacement field, (c) exposure-dependent blur, and (d) final synthesized turbulence combining both degradation components.}
  \label{fig:rows1to6_gap}
\end{figure*}

% 第二个 sidewaysfigure


\end{document}